\documentclass[10pt,twocolumn,letterpaper]{article}

\usepackage[pagenumbers]{wacv} % To force page numbers, e.g. for an arXiv version

\usepackage[accsupp]{axessibility}  % Improves PDF readability for those with disabilities.

\usepackage{multirow}
\usepackage{algorithm}
\usepackage{algorithmic}

\definecolor{wacvblue}{rgb}{0.21,0.49,0.74}
\usepackage[pagebackref,breaklinks,colorlinks,allcolors=wacvblue]{hyperref}

\usepackage{amssymb}
\usepackage{pifont}
\newcommand{\cmark}{\ding{51}}
\newcommand{\xmark}{\ding{55}}

\def\wacvPaperID{3237} % *** Enter the WACV Paper ID here
\def\confName{WACV}
\def\confYear{2027}

\newcommand{\model}{\texttt{{IECG}}\xspace}
\title{\model: Weakly-Supervised Interpretability-Enhancing Foundation Model for Electrocardiogram}

\author{
Huynh Dang Nguyen$^{1}$ \quad
Trong-Thang Pham$^{2}$ \quad
Anh-Thuan Do$^{3,\dagger}$ \quad
Phuong-Nhi Dinh$^{4,\dagger}$ \\
Brijesh Patel$^{5}$ \quad
Ngan Le$^{2,*}$ \quad
Van Nguyen$^{1,*}$
\vspace{0.3em} \\
$^{1}$Quantum AI \& Cyber Security Institute (QACI) - FPT Corporation, Vietnam \\
$^{2}$University of Arkansas, USA \qquad
$^{3}$Hospital Haiphong Medical University, Vietnam \\
$^{4}$Pham Ngoc Thach University of Medicine, Vietnam \qquad
$^{5}$Indiana University, USA \\
{\tt\small \{HuynhND11, VanNTH19\}@fpt.com, \{tp030, thile\}@uark.edu} \\
{\tt\small 2151010317@hpmu.edu.vn, 2051010735@pnt.edu.vn, patebr@iu.edu}
}

\begin{document}
\maketitle
\begingroup
\renewcommand\thefootnote{}
\footnotetext{$^\dagger$Equal contribution. \quad $^*$Equal advising.}
\endgroup
\begin{abstract}
Electrocardiography (ECG) is a cornerstone of cardiac diagnosis, yet most machine learning methods for ECG analysis lack interpretability, limiting their clinical adoption. We present \textbf{\model}, a weakly-supervised ECG foundation model that jointly performs diagnostic classification and abnormality localization without requiring manual abnormal region annotations, overcoming a long-standing bottleneck in interpretable ECG learning. Central to \model is a weakly-supervised attention strategy that explicitly incorporates clinical priors by deriving abnormality heatmap pseudo-labels from established ECG clinical knowledge. These pseudo-labels guide the model's CLS-token attention through a KL-divergence loss, encouraging it to focus on diagnostically relevant waveform regions. During pretraining, \model is jointly optimized with clinically guided attention supervision, ECG-text contrastive alignment, and report captioning objectives, enabling it to learn rich multimodal representations while producing localized waveform evidence. By explicitly aggregating features from clinically guided attention regions, \model grounds its diagnostic predictions in interpretable waveform evidence without sacrificing classification performance. Extensive experiments demonstrate that \model achieves state-of-the-art abnormal region localization while maintaining strong classification performance, advancing trustworthy foundation models for ECG analysis.
\end{abstract}
    
\section{Introduction}
\label{sec:intro}

Deep learning architectures for electrocardiogram (ECG) analysis~\cite{ecg-chat,tolerantecg,stmem,ecgfm} have achieved remarkable success in automated cardiac diagnosis. However, despite their strong predictive performance, these models remain largely uninterpretable: they predict \emph{what} abnormality is present but fail to explain \emph{where} the supporting evidence resides within the ECG waveform. This limitation substantially hinders their clinical adoption, as physicians rely not only on diagnostic predictions but also on verifying the underlying waveform morphology. An ECG analysis model should therefore identify both the predicted diagnosis and the localized waveform evidence supporting that prediction.

Despite its importance, abnormality localization remains an unsolved problem for ECG foundation models. The primary challenge is the lack of localized supervision. Large-scale ECG datasets such as MIMIC-IV-ECG~\cite{mimic-iv} and PTB-XL~\cite{ptb-xl} provide only recording-level diagnostic labels and contain no annotations identifying the waveform regions corresponding to each diagnosis. Consequently, existing models are optimized solely for classification and cannot learn to consistently associate predictions with clinically meaningful waveform regions. While various methods can highlight irregular beats \cite{bui2024tsrnetsimpleframeworkrealtime, jiang2023multiscalecrossrestorationframeworkelectrocardiogram}, they fail to provide specific diagnoses associated with those localized regions.

In this work, we address this challenge by exploiting a key observation: although localized annotations are unavailable, diagnostic criteria for many ECG abnormalities are well established in clinical practice. For example, a PR interval exceeding 200~ms is a diagnostic criterion for first-degree AV block, while ST-segment elevation greater than 0.1~mV indicates myocardial injury. We formalize these clinical heuristics into an automatic pseudo-label generation pipeline that produces abnormality heatmaps directly from ECG signals. By explicitly incorporating these clinical priors, we transform localization from a manual annotation problem into a weakly-supervised learning problem.

Building upon this insight, we propose \textbf{\model}, an attention-enhanced interpretable ECG foundation model that jointly performs classification and abnormality localization. During training, clinically derived abnormality heatmaps supervise the model's attention through a Kullback-Leibler (KL) divergence objective, encouraging attention to align with diagnostically relevant waveform regions while learning disease classification. As a result, \model produces both diagnostic predictions and localized visual evidence supporting those predictions. For example, when diagnosing atrial fibrillation, the model highlights the irregularly absent P-wave regions that cardiologists naturally inspect during clinical interpretation.

Extensive experiments demonstrate that \model achieves state-of-the-art (SOTA) abnormal region localization, reaching 75.55\% AUROC and outperforming existing attribution methods, including approaches specifically designed for this task, while archiving competitive classification performance. These findings suggest that clinically guided attention supervision can significantly enhance the interpretability of ECG foundation models without requiring manual attention annotations.

Our main contributions are summarized as follows:
\begin{itemize}
\item We propose \textbf{\model}, an ECG foundation model that jointly performs classification and abnormality localization without requiring manual attention annotations.
\item We introduce a weakly-supervised attention supervision framework that incorporates clinical priors to automatically generate abnormality heatmap pseudo-labels from established ECG diagnostic heuristics, aligning model attention with clinically meaningful waveform regions.
\item We demonstrate SOTA abnormality localization and competitive classification performance, highlighting the effectiveness of clinically guided attention supervision for improving the interpretability of ECG foundation models.
\end{itemize}
\section{Related Work}
\subsection{Foundation models for ECG}
Foundation models have become a dominant approach for ECG analysis, typically pretrained on large-scale datasets and adapted to downstream tasks through transfer learning \cite{ecgfm, any-ecg, crema, ecg-ssl}. However, existing approaches focus primarily on representation learning and predictive performance without explicitly addressing interpretability.

Recent work has explored both signal-only and multimodal pretraining strategies. METS~\cite{mets}, MERL~\cite{merl}, and TolerantECG~\cite{tolerantecg} adopt multimodal alignment to learn joint representations between ECG signals and diagnostic reports. These methods improve semantic representation quality but do not explicitly localize diagnostic evidence within the waveform. In contrast, signal-only self-supervised methods such as ST-MEM~\cite{stmem}, ECG-FM~\cite{ecgfm}, and HeartLang~\cite{readingheartlearningecg} leverage masked reconstruction or contrastive learning to capture intrinsic spatio-temporal representations of ECG signals.

\textit{Existing ECG foundation models focus on representation, yet they do not explicitly solve where in the waveform the evidence for a diagnosis lies. In contrast, \model is designed to enhance interpretability by construction: it introduces CGAL to generate attention pseudo-labels from clinical heuristics and uses these labels to directly supervise the model’s CLS-token attention via a KL-divergence loss, enabling anomaly localization without manual attention annotation while preserving strong downstream performance.}

\subsection{Interpretable AI and Localization}
Explainable AI aims to enhance transparency in deep learning systems, which is particularly important in the medical domain, where clinicians must understand a model’s reasoning before trusting its predictions. 

Saliency-based methods~\cite{saliency, xai4ecg} compute gradients of the model output with respect to the input to estimate feature importance. While these approaches can highlight influential regions, they often produce noisy and unstable explanations. In contrast, Grad-CAM~\cite{grad-cam, ecg_xplaim} generates coarse-resolution heatmaps by leveraging gradients and intermediate feature maps. However, in the context of ECG signals, such heatmaps primarily capture temporal importance and do not explicitly disentangle contributions from individual leads, which limits their clinical interpretability.

For Transformer architectures, attention calibration \cite{lu-etal-2021-attention, zhou2024attentioncalibrationtransformerbasedsequential} introduces auxiliary objectives to guide attention distributions toward more meaningful and interpretable patterns. Existing approaches typically derive such supervision signals from model-internal mechanisms, such as perturbation-based importance estimation or structural constraints, rather than relying on explicit ground-truth attention annotations. 

\textit{However, these generic calibration strategies are not tailored to domain-specific interpretability requirements. In the ECG domain, clinically meaningful attention should align with abnormal waveform regions, yet such supervision signals are not readily available in existing datasets. Our work addresses this gap by automatically generating pseudo-labels from clinical heuristics, enabling domain-specific attention calibration without manual labeling.}

\section{Model Architecture}

\begin{figure}[t]
  \centering
  \includegraphics[width=1.0\columnwidth]{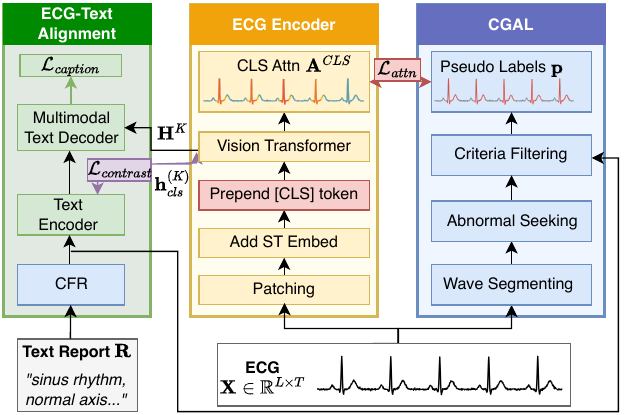}
   \caption{\model overview. The ECG Encoder learns representations through ECG-Text Alignment and a self-supervised clinical abnormality localization mechanism with CGAL pipeline.}
   \label{fig:model_overview}
\end{figure}

As illustrated in Figure~\ref{fig:model_overview}, \model consists of three main components: the ECG Encoder, ECG-Text Alignment, and CGAL. Given an ECG signal $\mathbf{X} \in \mathbb{R}^{L \times T}$ and its corresponding text report $\mathbf{R}$, \model learns multimodal representations while simultaneously learning to focus on clinically relevant regions.

The ECG Encoder (center) processes the raw 12-lead ECG through patching, Spatial-Temporal (ST) embedding (Section~\ref{sec:st-embed}), and a Vision Transformer (ViT) to produce ECG representations. The CLS token aggregates global information, and its attention scores over the ECG patches form the basis for interpretability. ECG-Text Alignment (left) aligns ECG representations with text reports preprocessed by the Criteria Feature Retrieval (CFR) module, using a frozen Text Encoder and a Multimodal Text Decoder optimized through contrastive ($\mathcal{L}_{contrast}$) and captioning ($\mathcal{L}_{caption}$) losses (Section~\ref{sec:alignment}). CGAL (right) addresses the annotation bottleneck by automatically generating pseudo-labeled abnormal heatmaps. The process begins by segmenting ECG waveforms, after which the Abnormal Seeking module identifies irregular regions using clinical heuristics. Subsequently, the Criteria Filtering module refines these regions into pseudo labels based on clinical criteria retrieved by CFR. The attention loss ($\mathcal{L}_{attn}$) then guides the CLS attention toward these clinically meaningful regions (Section~\ref{sec:cgal}).

\subsection{Spatial-Temporal Embedding} \label{sec:st-embed}

Given a raw 12-lead ECG signal $\mathbf{X} \in \mathbb{R}^{L \times T}$, where $L=12$ denotes the number of leads and $T=5000$ is the temporal length (10 seconds at 500Hz), we first apply a linear patch embedding layer. Each lead is divided into $P = T/s = 100$ non-overlapping patches of size $s=50$, yielding patch embeddings $\mathbf{Z}^{(l)} \in \mathbb{R}^{P \times D}$ for each lead $l \in \{0, \ldots, L-1\}$, where $D$ is the hidden dimension. We then flatten across leads by concatenating all patches sequentially (I, II, III, aVR, aVL, aVF, V1, V2, V3, V4, V5, V6):
\begin{equation}
    \mathbf{Z} = [\mathbf{Z}^{(0)}; \mathbf{Z}^{(1)}; \ldots; \mathbf{Z}^{(L-1)}] \in \mathbb{R}^{N \times D},
\end{equation}
where $N = L \times P = 1200$ is the number of tokens.

\begin{figure}[t]
  \centering
   \includegraphics[width=1.0\linewidth]{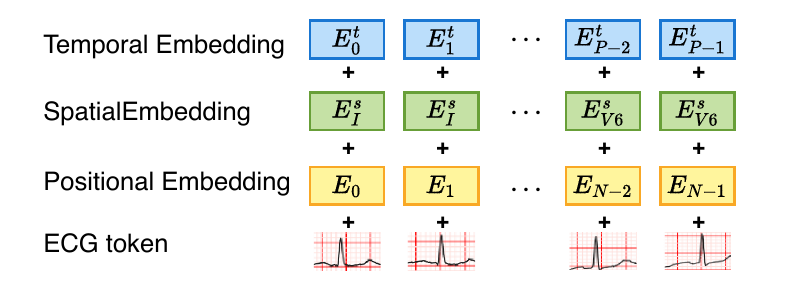}
    \caption{Each ECG token is enriched with Spatial and Temporal embeddings in addition to Positional embeddings, providing explicit lead identity and time frame information.\vspace{-.5cm}}
    \label{fig:st-embed}
\end{figure}

In addition to the Transformer's positional embedding $\mathbf{E} \in \mathbb{R}^{N \times D}$, we incorporate the Spatial-Temporal (ST) embedding proposed by \citet{readingheartlearningecg} to enhance the model's ability to capture lead-specific and temporal information, as illustrated in Figure~\ref{fig:st-embed}. For spatial embedding, we define $\mathbf{E}^{s} \in \mathbb{R}^{L \times D}$, assigning a unique embedding to each lead (e.g., $E^s_{\text{I}}, E^s_{\text{II}}, \ldots, E^s_{\text{V6}}$). For temporal embedding, we define $\mathbf{E}^{t} \in \mathbb{R}^{P \times D}$, encoding the position within each lead (i.e., $E^t_0, E^t_1, \ldots, E^t_{P-1}$). The input to the Transformer is:
\begin{equation}
\mathbf{H}^{(0)}_n = \mathbf{Z}_n + \mathbf{E}_n + \mathbf{E}^{s}_{l(n)} + \mathbf{E}^{t}_{t(n)}
\end{equation}
where $l(n) = \lfloor n/P \rfloor$ and $t(n) = n \bmod P$ for $n = 0, \ldots, N-1$. Here $\mathbf{H}^{(0)}_n \in \mathbb{R}^{D}$ is the $n$-th token's input representation (0-indexed), combining its patch embedding with positional embedding ($E_0, E_1, \ldots, E_{N-1}$), spatial embedding, and temporal embedding.

\subsection{ECG Signal and Text Report Alignment} \label{sec:alignment}

Once the ECG tokens are enriched with spatial-temporal information, they are passed through the ECG Encoder, a standard Vision Transformer with $K$ layers. A learnable CLS token $\mathbf{h}_{cls}^{(0)} \in \mathbb{R}^{D}$ is prepended to the sequence:
\begin{equation}
    \tilde{\mathbf{H}}^{(0)} = [\mathbf{h}_{cls}^{(0)}; \mathbf{H}^{(0)}] \in \mathbb{R}^{(N+1) \times D}.
\end{equation}
The encoder applies $K$ transformer layers with multi-head self-attention (MHSA) and feed-forward networks (FFN):
\begin{equation}
    \tilde{\mathbf{H}}^{(k)} = \text{FFN}(\text{MHSA}(\tilde{\mathbf{H}}^{(k-1)})), \quad k = 1, \ldots, K.
\end{equation}
The final CLS token $\mathbf{h}_{cls}^{(K)} \in \mathbb{R}^{D}$ serves as the global ECG representation $\mathbf{E}$ for downstream tasks.

Following previous ECG pretraining approaches \cite{esi, ecg-chat}, we employ two pretraining objectives: contrastive loss for representation learning and captioning loss for semantic alignment. Before encoding, text reports are processed by the Criteria Feature Retrieval (CFR) module~\cite{tolerantecg}, which retrieves relevant diagnostic criteria to enrich each diagnosis with their corresponding features. Following CoCa~\cite{coca}, the Multimodal Text Decoder employs causal self-attention and cross-attention over ECG encoder outputs for autoregressive report generation. The ECG Encoder and Multimodal Text Decoder are jointly optimized while the Text Encoder remains frozen.

Optimizing the contrastive loss enables alignment between ECG features $\mathbf{E} \in \mathbb{R}^{D}$ and corresponding text embeddings $\mathbf{T} \in \mathbb{R}^{D}$. For a batch of $B$ samples, let $\{\mathbf{E}_i, \mathbf{T}_i\}_{i=1}^{B}$ denote the paired ECG and text representations, and $\tau$ the temperature parameter:
\begin{equation}
\begin{split}
    \mathcal{L}_{contrast} &= -\frac{1}{2B} \Bigg(
        \sum_{i=1}^{B} \log \frac{e^{\mathbf{E}_i^\top \mathbf{T}_i / \tau}}
        {\sum_{j=1}^{B} e^{\mathbf{E}_i^\top \mathbf{T}_j / \tau}}
        \\ &+ 
        \sum_{i=1}^{B} \log \frac{e^{\mathbf{T}_i^\top \mathbf{E}_i / \tau}}
        {\sum_{j=1}^{B} e^{\mathbf{T}_i^\top \mathbf{E}_j / \tau}}
    \Bigg)
\end{split}
\label{eq:contrastive}
\end{equation}

Optimizing the captioning loss encourages the ECG encoder to produce representations that enable accurate report generation. Given the ECG token sequence ${\mathbf{H}}^{(K)} \in \mathbb{R}^{(N) \times D}$ from the encoder (excluding CLS token) and target text tokens $\mathbf{t} = (t_1, \ldots, t_M)$ of length $M$, the Multimodal Text Decoder with parameters $\theta$ autoregressively predicts each token, with the corresponding loss function:
\begin{equation}
    \mathcal{L}_{caption} = - \mathbb{E} \Bigg [\sum_{m=1}^{M}\log P_\theta(t_m | t_{1:m-1}, {\mathbf{H}}^{(K)})\Bigg ]
    \label{eq:captioning}
\end{equation}

\subsection{Clinically-Guided Attention Localization (CGAL)} 
\label{sec:cgal}

CGAL is the key component that solves the annotation bottleneck. It generates attention pseudo-labels during training by encoding established clinical knowledge into a rule-based algorithm. The overall procedure is detailed in Section~\ref{sec:appendix_cgal} of the Supplementary Material.

The first stage, \textit{Wave Segmenting}, focuses on segmenting the ECG waveform into its constituent components. We begin by denoising the ECG signal through successive application of a high‑pass filter, notch filter, and low‑pass filter. A pretrained UNet3+ model~\cite{unet3p, Joung2024DeepECGSegmentation}, trained on SemiSegECG~\cite{semisegecg} and RDB~\cite{rest-ecg} datasets and frozen during our training, produces a segmentation map $\mathbf{S} \in \{0,1,2,3\}^{T}$, where labels correspond to background (0), P wave (1), QRS complex (2), and T wave (3). The detailed performance of this segmentation module is reported in Section~\ref{sec:unet_performance} of the Supplementary material.

The second stage, \textit{Abnormal Seeking}, identifies candidate abnormal regions using clinical definitions from the \textit{Life in the Fastlane} (LITFL) ECG library~\cite{litfl}. Firstly, when P wave is not visible (e.g., in atrial fibrillation), we insert a pseudo P wave between the preceding T wave and the current Q wave, and highlight this region to indicate the absence of a true P wave. Based on the segmented waveform boundaries, clinically relevant segments and intervals are computed, including the PR interval, defined as the interval from P wave onset to QRS onset, as well as the QT interval and the ST segment. For QT interval analysis, we apply Bazett’s correction to ensure clinical validity: $QTc = QT / \sqrt{RR}$, where $QT$ is the measured interval and $RR$ is the preceding R-R interval. This correction accounts for the dependence of the QT interval on heart rate. Finally, regions exceeding clinically established amplitude and duration thresholds, as defined in Section~\ref{sec:abnormal_threshold} of the Supplementary material, are identified and flagged as potential abnormalities.

\begin{figure}[t]
  \centering
    \includegraphics[width=1.0\linewidth]{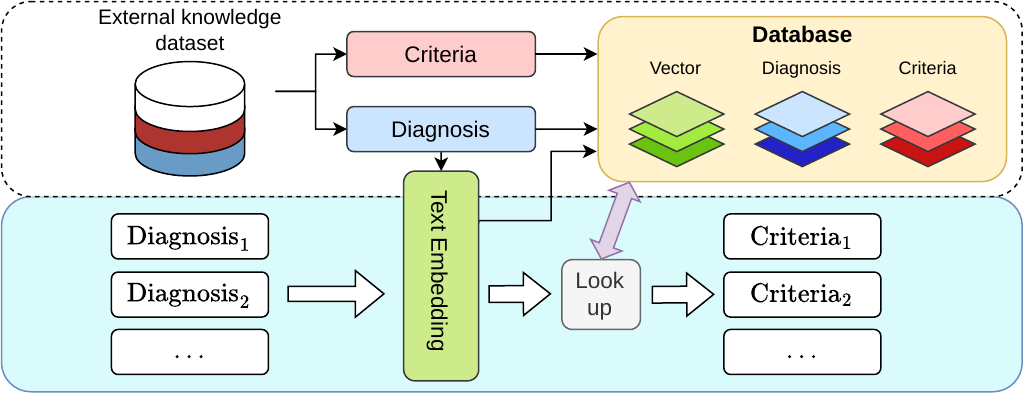}
    \caption{The Criteria Feature Retrieval (CFR) module retrieves diagnostic criteria by comparing the input diagnosis embedding with a pre-constructed vector database.} 
    \vspace{-.5cm}
    \label{fig:cfr}    
\end{figure}

The third stage, \textit{Criteria Filtering}, filters these candidates using the CFR module~\cite{tolerantecg}, illustrated in Figure~\ref{fig:cfr}. CFR maintains a vector database $\mathcal{D} = \{(\mathbf{d}_j, c_j)\}_{j=1}^{|\mathcal{D}|}$ of diagnosis-criteria pairs sourced from the LITFL ECG library~\cite{litfl}, where $\mathbf{d}_j \in \mathbb{R}^{D}$ is the embedding of diagnosis $j$ and $c_j$ is its defining criteria (e.g., ``prolonged PR interval $>$ 200ms'' for first-degree AV block). Given an input diagnosis with embedding $\mathbf{q} \in \mathbb{R}^{D}$, the CFR module retrieves the most similar criteria via cosine similarity.

We retain only those flagged regions that match the retrieved criteria. For example, regions corresponding to prolonged PR intervals are preserved for first degree atrioventricular block, while unrelated findings such as ST segment abnormalities are discarded. When no matching criteria can be retrieved, a fallback procedure is applied, which uses the input query itself to identify candidate abnormal regions, as detailed in Section~\ref{sec:cfr_quality} of the Supplementary material.

The selected regions are then converted into a binary token mask $\mathbf{m} \in \{0, 1\}^{L \cdot T}$, where a value of 1 indicates that the corresponding token lies within a flagged region in $\mathcal{R}$, and a value of 0 denotes all other positions. When multiple diagnoses are associated with a signal, the final mask is obtained by applying a logical OR operation across the diagnosis-specific masks.

\textbf{Attention Supervision.} From the preceding stages, we obtain a binary mask $\mathbf{m}_i \in \{0, 1\}^{L \cdot T}$ for sample $i$, where $m_{i,k} = 1$ indicates an abnormal region at timestamp $k$. However, since the attention token got downsampled after proccessed by ECG encoder, we apply a MaxPool operation with kernel size and stride same as patch size $s$. We further normalize this mask into a target probability distribution $\mathbf{p}_i \in \mathbb{R}^{N}$:
\begin{align}
\resizebox{0.9\columnwidth}{!}{
    $\tilde{\mathbf{m}_i} = \text{MaxPool}(\mathbf{m}_i, \text{kernel\_size}=s, \text{stride}=s) \in \mathbb{R}^N$
}\\
\resizebox{0.8\columnwidth}{!}{
    $\mathbf{p}_{i,n} = \tilde{\mathbf{m}}_{i,n} / \sum_{j=0}^{N-1} \tilde{\mathbf{m}}_{i,j}, \quad n = 0, \ldots, N-1$
}
\label{eq:pseudo_label}
\end{align}

To extract the model's global attention pattern, we compute the aggregated Attention Rollout across all transformer layers \cite{attn-rollout}. This technique accounts for the flow of information through residual connections and successive self-attention operations. For a given layer $k \in \{1, \ldots, K\}$, let $\mathbf{W}^{(k)} \in \mathbb{R}^{(N+1) \times (N+1)}$ denote the attention matrix averaged across all $H$ heads:
\begin{equation}
    \mathbf{W}^{(k)} = \frac{1}{H} \sum_{h=1}^{H} \text{softmax}\left(\frac{\mathbf{Q}^{(k,h)} {\mathbf{K}^{(k,h)}}^\top}{\sqrt{d_h}}\right)
\end{equation}
To incorporate residual connections, the attention weights are augmented with an identity matrix and re-normalize \cite{attn-rollout}: 
% Quote from author: "Given the attention module with residual connection, we compute values in layer $l+1$ as $V_{l+1} = V_l + W_{att} V_l$, where W_{att} is the attention matrix. Thus, we have $V_{l+1} = (W_{att} + I) V_l$. So, to account for residual connections, we add an identity matrix to the attention matrix and re-normalize the weights. This results in $A = 0.5 W_{att} + 0.5 I$, where A is the raw attention updated by residual connections."
%
\begin{equation}
    \tilde{\mathbf{W}}^{(k)} = (\mathbf{W}^{(k)} + \mathbf{I}) / 2
\end{equation} 
The rollout attention matrix $\mathbf{R}^{(k)}$ is computed recursively by multiplying the augmented attention matrices from the current layer with the rollout matrix from the previous layer:
\begin{equation}
    \mathbf{R}^{(k)} = \tilde{\mathbf{W}}^{(k)} \mathbf{R}^{(k-1)}
\end{equation}
where $\mathbf{R}^{(0)} = \mathbf{I}$. The final aggregated attention scores from the CLS token to the $N$ ECG tokens are extracted from the CLS-token row of the final rollout matrix $\mathbf{R}^{(K)}$, denoted as $\mathbf{A}_i^{rollout} \in \mathbb{R}^{N}$. The attention loss aligns this aggregated distribution with the pseudo-label via KL divergence:
\begin{equation}
    \mathcal{L}_{attn} = \mathbb{E} \Bigg [\sum_{i=1}^B \sum_{n=0}^{N-1} {\mathbf{p}_{i,n}} \log ({\mathbf{p}_{i,n}} / \mathbf{A}_{i,n}^{rollout})\Bigg ]
\label{eq:attn_loss}
\end{equation}

By utilizing attention rollout, the supervision signal explicitly guides the hierarchical information aggregation process of the entire Transformer, rather than constraining only the final block. This design ensures that intermediate layers progressively learn to filter and route relevant morphological features toward the CLS token, leading to more robust and coherent localization. 

To further evaluate the effectiveness of this design, we additionally consider a variant in which attention supervision is applied only to the CLS-token attention scores from the final Transformer layer.

\subsection{Training Objective}
The overall training loss combines all objectives:
\begin{equation}
    \mathcal{L}_{total} = \alpha \mathcal{L}_{contrast} + \beta \mathcal{L}_{caption} + \gamma \mathcal{L}_{attn}
    %\label{total_loss}
\end{equation}
where $\alpha$, $\beta$ and $\gamma$ control the relative contributions of contrastive learning, captioning, and attention supervision, respectively. For samples where a valid heatmap cannot be constructed by CGAL, attention supervision is omitted, and training proceeds using only $\mathcal{L}_{\text{contrast}}$ and $\mathcal{L}_{\text{caption}}$.

Through this composite objective, contrastive loss aligns ECG features with textual representations, captioning loss enables accurate generation of clinical descriptions, and attention-guiding loss regularizes the model to focus on medically relevant regions.

\subsection{Down-stream Evaluation}

After pretraining, the ECG encoder can be deployed for a variety of downstream tasks. One direct application is zero‑shot abnormal region localization, achieved by collecting the output attention scores from input signal. CGAL is omitted during inference since the model has already learned to localize relevant regions. In addition, by attaching a linear classification layer on top of the encoder, the model can be further fine‑tuned for classification. Importantly, since only the classifier head is updated while the encoder remains frozen (linear probing), the ability to attend to clinically contributing regions is preserved. 
\section{Experiment}
\subsection{Datasets}
We pretrain on MIMIC-IV-ECG \cite{mimic-iv}, comprising over 800,000 12-lead ECG recordings (10-second duration) from approximately 160,000 patients. During preprocessing, we replace NaN/Inf values with zero. CGAL generates attention pseudo-labels without external annotations.

For abnormality localization, we benchmark using the dataset proposed by \citet{ecgad}, which we refer to as $\text{ECGAD}_{\text{data}}$, consistent with the model’s name $\text{ECGAD}_{\text{model}}$. The dataset consists of 400 PTB-XL test samples annotated by two cardiologists. Among these, only 190 samples are abnormal and include heatmap annotations. We use these annotated cases for benchmarking. 

For classification evaluation, since $\text{ECGAD}_{\text{data}}$ is a subset of PTB-XL~\cite{ptb-xl}, we utilize the corresponding labels of these samples for this task. In addition, we evaluate on the full PTB-XL dataset, a clinical 12-lead ECG corpus comprising 21,799 recordings from 18,869 patients, using the predefined train, validation, and test splits. We benchmark on All-level, which includes all 71 diagnostic categories. 

\subsection{Configuration}
We use a 1D 12-layer Vision Transformer with 12 attention heads, a hidden width of 768, and a patch size of 50 as the ECG encoder. The text encoder (MedCPT \cite{jin2023medcpt}) remains frozen. We flatten the 12-lead ECG signal to ensure the CLS token attends to all tokens. Training uses learning rate $1e-4$, 20 epochs, batch size 32 on $4\times$ A100 GPUs. We set $\alpha = 1$, $\beta = 1$ (following \cite{esi}), and $\gamma=0.5$. 

\subsection{Metrics}
\textbf{Abnormality localization.} To evaluate whether the attention scores of \model accurately highlight relevant regions within each data sample, we frame this as a ranking task where the model should assign higher attention to abnormal tokens. We employ Precision at K (P@K) ($\uparrow$) where K = 50\% of the total token count (K=30,000 for $L \times T = 60,000$), as the large sequence length makes fixed small K values less discriminative. We also report point‑level Area Under the Receiver Operating Characteristic Curve (AUROC) ($\uparrow$), following the protocol in \cite{ecgad}.

\noindent\textbf{Classification.}: We evaluate classification performance using FMax ($\uparrow$), the maximum F1 score over all thresholds \cite{fmax}, summarizing precision–recall balance. We also report macro-averaged AUROC ($\uparrow$), which measures overall discriminative capability across all classes regardless of class imbalance. For clarity, we denote localization and classification AUROC as AUROC$_{\text{loc}}$ and AUROC$_{\text{cls}}$, respectively.

\noindent\textbf{Faithfulness.}: We evaluate the faithfulness of the generated attention maps using Insertion ($\uparrow$) and Deletion ($\downarrow$) tests \cite{insert_delete}. Specifically, we measure AUROC after progressively inserting or deleting the top-\(k\) attended regions, where \(k \in \{10\%, 20\%, \ldots, 100\%\}\).

\begin{table*}
\centering
\caption{Abnormality localization and Classification results on $\text{ECGAD}_{\text{data}}$. \textbf{Bold} indicates the best result, and \underline{underline} denotes the second-best result. For \model, attention supervision is applied to either the attention scores from the final Transformer layer (\model-Last) or the layer-wise aggregated attention obtained via attention rollout (\model-Rollout).
% \textcolor{red}{add binary anomal classification.}\textcolor{red}{consider add the best classification.} \textcolor{red}{use only insertion and detection. Check the calculation.}
}
\label{tab:main_table}
\resizebox{0.9\textwidth}{!}{
\begin{tabular}{l|c|c|c|c|c|c|c}
    \bf Model & \bf Interpretable & \multicolumn{2}{c|}{\bf Abnormality localization} & \multicolumn{2}{c|}{\bf Classification} & \multicolumn{2}{c}{\bf Faithfulness} \\
    &  & \bf P@50\% ($\uparrow$) & \bf AUROC$_{\text{loc}}$ ($\uparrow$) & \bf FMax ($\uparrow$) & \bf AUROC$_{\text{cls}}$ ($\uparrow$) & Insertion ($\uparrow$) & Deletion ($\downarrow$) \\
    \midrule
    ST-MEM \cite{stmem} & \xmark & 16.56 \tiny{$\pm$1.29} & 50.07 \tiny{$\pm$0.65} & 16.18 \tiny{$\pm$0.12} & 59.07 \tiny{$\pm$0.25} & 23.25 \tiny{$\pm$0.38} & 21.57 \tiny{$\pm$0.37} \\
    METS \cite{mets} & \xmark & 18.89 \tiny{$\pm$1.26} & 59.25 \tiny{$\pm$1.24} & 40.19 \tiny{$\pm$0.33} & 88.05 \tiny{$\pm$0.30} & 38.42 \tiny{$\pm$0.58} & 21.24 \tiny{$\pm$0.48} \\
    MERL \cite{merl} & \xmark & 19.73 \tiny{$\pm$1.28} & 61.13 \tiny{$\pm$1.23} & 41.82 \tiny{$\pm$0.57} & 88.80 \tiny{$\pm$0.27} & 35.50 \tiny{$\pm$0.28} & \underline{20.07 \tiny{$\pm$0.54}} \\
    ECG-FM \cite{ecgfm} & \xmark & 16.27 \tiny{$\pm$1.21} & 50.29 \tiny{$\pm$0.43} & 34.37 \tiny{$\pm$0.43} & 83.41 \tiny{$\pm$0.52} & 25.23 \tiny{$\pm$0.36} & \textbf{12.09 \tiny{$\pm$0.14}} \\
    HeartLang \cite{readingheartlearningecg} & \xmark & 17.08 \tiny{$\pm$1.28} & 50.32 \tiny{$\pm$1.12} & 27.58 \tiny{$\pm$0.47} & 77.07 \tiny{$\pm$0.64} & 23.50 \tiny{$\pm$0.31} & 20.76 \tiny{$\pm$0.34} \\
    TolerantECG \cite{tolerantecg} & \xmark & 18.25 \tiny{$\pm$1.22} & 59.15 \tiny{$\pm$1.21} & \underline{42.26 \tiny{$\pm$0.22}} & \textbf{90.26 \tiny{$\pm$0.30}} & 35.89 \tiny{$\pm$0.86} & 22.75 \tiny{$\pm$1.02} \\
    %\midrule
    XAI4ECG \cite{xai4ecg} & \cmark & 16.63 \tiny{$\pm$0.08} & 51.44 \tiny{$\pm$0.32} & 40.21 \tiny{$\pm$1.09} & 88.02 \tiny{$\pm$0.44} & 27.39 \tiny{$\pm$0.71} & 25.43 \tiny{$\pm$0.54} \\
    ECG‑XPLAIM \cite{ecg_xplaim} & \cmark & 17.43 \tiny{$\pm$0.25} & 52.56 \tiny{$\pm$0.93} & -- & -- & -- & -- \\
    %\midrule
    $\text{ECGAD}_{\text{model}}$ \cite{ecgad} & \cmark & 22.28 \tiny{$\pm$1.37} & 71.62 \tiny{$\pm$1.21} & 28.30 \tiny{$\pm$0.69} & 75.01 \tiny{$\pm$1.28} & 32.62 \tiny{$\pm$1.36} & 22.70 \tiny{$\pm$0.65} \\
    \midrule
    {\model}-Last & \cmark & \underline{23.67 \tiny{$\pm$1.48}} & \underline{73.94 \tiny{$\pm$0.99}} & \textbf{42.41 \tiny{$\pm$0.13}} & \underline{90.03 \tiny{$\pm$0.27}} & \underline{43.27 \tiny{$\pm$0.20}} & 26.18 \tiny{$\pm$0.42} \\
    
    {\model}-Rollout & \cmark & \textbf{24.42 \tiny{$\pm$1.48}} & \textbf{75.55 \tiny{$\pm$1.00}} & 42.00 \tiny{$\pm$0.43} & 89.09 \tiny{$\pm$0.49} & \textbf{45.86 \tiny{$\pm$0.55}} & 33.15 \tiny{$\pm$1.15} \\
\end{tabular}
}
\end{table*}

\subsection{Baselines}

We compare \model against a diverse set of baseline methods that cover both abnormality localization and classification.

To ensure a fair comparison, we retrain METS~\cite{mets}, MERL~\cite{merl}, and TolerantECG~\cite{tolerantecg} using the same encoder architecture and training dataset as \model. For ECG-FM~\cite{ecgfm} and HeartLang~\cite{readingheartlearningecg}, we use the official pretrained checkpoint. In addition, since ST-MEM~\cite{stmem} was not originally trained on the MIMIC-IV-ECG dataset~\cite{mimic-iv}, we retrain it to match our configuration.

During evaluation, ST-MEM and ECG-FM do not include a CLS token. Following the design choices in their original implementations, where token representations are aggregated using average pooling before being passed to a classification head. We adopt the same aggregation strategy to derive the corresponding attention scores. In contrast, because METS, MERL, HeartLang, and TolerantECG share the same architecture as our model, their attention scores are extracted using the same procedure as ours.

For XAI4ECG~\cite{xai4ecg} and ECG-XPLAIM~\cite{ecg_xplaim}, which generate attribution maps for interpretability, we retrain the models following their released implementations. However, ECG-XPLAIM is trained in a task-specific supervised setting and is designed for a limited set of classification tasks rather than the full multi-label All-level setting in PTB-XL. As a result, its training protocol is not directly compatible with our classification benchmark; hence classification and faithfulness results are not reported for this model.

In contrast, $\text{ECGAD}_{\text{model}}$~\cite{ecgad} is specifically designed for abnormality localization, and we use the publicly available checkpoint weights. Following \cite{jiang2024selfsupervisedanomalydetectionpretraining}, the latent representations produced by $\text{ECGAD}_{\text{model}}$ are passed through a ResNet classifier for downstream classification. 
% Consequently, abnormality localization and classification are performed by separate components rather than through a unified end-to-end pipeline as in our approach.

In all cases, we directly use the attribution or localization maps produced by each method as attention scores for evaluation.

\section{Results and Analysis} \label{sec:results}

\subsection{Localization and Classification Results}
\begin{figure*}[t]
  \centering
    \includegraphics[width=0.9\linewidth]{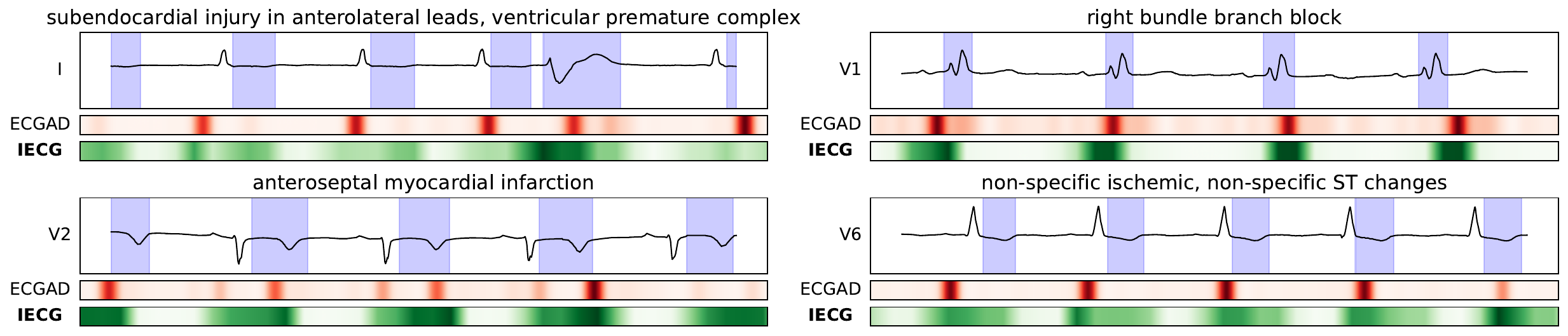}
    \caption{Qualitative comparison of abnormality localization across different ECG conditions. For each example, the top panel shows the ECG waveform with ground-truth abnormalities highlighted in blue. The middle and bottom rows present the localization heatmaps produced by ECGAD$_\text{model}$ and \model, respectively. Compared to ECGAD$_\text{model}$, \model produces more accurate and temporally consistent activations that better align with clinically relevant abnormalities across diverse conditions.
    }
    \label{fig:qualitative_compare}
\end{figure*}
To evaluate the model’s ability to identify abnormalities, the pretrained \model is directly applied to $\text{ECGAD}_{\text{data}}$ for inference, and the task is formulated as a binary pointwise classification problem~\cite{ecgad}. The ground truth labels are derived from expert annotations, which ensures that the evaluation remains independent of the pseudo labels used during training. To align the predicted attention heatmaps with the ECG input and the corresponding labels, we upsample the heatmaps using interpolation prior to metric computation.

For classification, since the samples in this dataset are derived from PTB-XL, we retrieve the corresponding labels for each sample. We then perform classification by appending a linear classification head to each model and training only this head on the PTB-XL training split, while keeping the backbone parameters frozen. The trained classifiers are subsequently evaluated on the $\text{ECGAD}_{\text{data}}$ samples.

Table~\ref{tab:main_table} summarizes the performance of all methods on both abnormality localization and classification tasks, as well as explanation faithfulness. Overall, both \model variants achieve the best localization performance across all metrics while obtaining highly competitive classification and faithfulness results, demonstrating its ability to jointly optimize predictive accuracy and clinically meaningful interpretability.

In terms of abnormality localization, both variants of \model consistently outperform all baseline methods by a clear margin across P@50\% and AUROC$_{\text{loc}}$. In particular, \model-Rollout achieves the best AUROC$_{\text{loc}}$ of 75.55\%, exceeding $\text{ECGAD}_{\text{model}}$ (71.62\%), which is specifically designed for localization, by a notable margin. This result highlights the effectiveness of incorporating CGAL during training, enabling the model to focus on diagnostically relevant waveform regions without requiring manual annotations. The improvement in P@50\% further indicates that \model ranks abnormalities more accurately, which is important for clinical interpretability.

In contrast, foundation models such as METS, MERL, and ECG-FM exhibit localization performance that is closer to random, with AUROC$_{\text{loc}}$ ranging from 50\% to 61\%. This suggests that optimizing solely for classification does not guarantee meaningful attention behavior. Although explainability-focused methods such as XAI4ECG and ECG-XPLAIM provide moderate improvements over some base models, their performance remains substantially below that of \model, indicating that post hoc explanation methods may lack precise alignment with clinically relevant regions.

As for classification, \model-Last achieves the highest FMax of 42.41\%, while TolerantECG attains the highest AUROC$_{\text{cls}}$ of 90.26\%, with \model-Last being highly competitive at 90.03\%. This marginal gap suggests that introducing attention supervision does not significantly compromise classification performance. Instead, \model preserves strong discriminative capability while substantially improving interpretability. In terms of explanation faithfulness, \model-Rollout attains the highest Insertion score of 45.86\%, confirming that the highlighted regions are indeed crucial for the model's predictions, though some baselines report lower Deletion scores.

Overall, these results demonstrate that explicitly guiding attention using clinically informed pseudo labels produced by CGAL allows \model to achieve a strong balance between localization, classification, and faithfulness. The proposed approach not only maintains state-of-the-art classification performance but also significantly improves the reliability and interpretability of the model’s predictions.

\begin{table}
    \centering
    \caption{Classification result evaluated on PTB-XL dataset. \textbf{Bold} indicates the best result, and \underline{underline} indicates the second-best result.
    % \textcolor{red}{Update the new results and move this table to main}
    }
    \label{tab:ptb-classify}
    \resizebox{.7\columnwidth}{!}{
        \begin{tabular}{l|c|c}
        \bf Model & \bf FMax & \bf AUROC$_{\text{cls}}$ \\
        \midrule
        ST-MEM \cite{stmem} & 11.05 \tiny{$\pm$ 0.03}  & 64.39 \tiny{$\pm$0.04} \\
        METS \cite{mets} & 39.88 \tiny{$\pm$0.23} & 90.82 \tiny{$\pm$0.17} \\
        MERL \cite{merl} & 40.00 \tiny{$\pm$0.27} & 91.66 \tiny{$\pm$0.17} \\
        ECG-FM \cite{ecgfm} & 31.00 \tiny{$\pm$0.33} & 85.10 \tiny{$\pm$0.32} \\
        HeartLang \cite{jin2025readingheartlearningecg} & 26.03 \tiny{$\pm$0.47} & 80.10 \tiny{$\pm$0.34} \\
        TolerantECG \cite{tolerantecg} & 40.76 \tiny{$\pm$0.38} & 92.02 \tiny{$\pm$0.12} \\
        \midrule
        \model-Last & \textbf{42.43 \tiny{$\pm$0.14}} & \textbf{92.21 \tiny{$\pm$0.14}} \\
        \model-Rollout & \underline{41.62 \tiny{$\pm$0.38}} & \underline{92.04 \tiny{$\pm$0.28}} \\ \bottomrule
        \end{tabular}
    }
\end{table}

For additional classification evaluation on the full PTB-XL dataset, Table~\ref{tab:ptb-classify} presents the classification results on the PTB-XL dataset. Overall, \model achieves the best performance across all metrics, demonstrating strong generalization when evaluated on large-scale ECG data.

Specifically, \model-Last achieves the best performance across both evaluation metrics, attaining the highest FMax of 42.43\% and AUROC$_{\text{cls}}$ of 92.21\%. Notably, \model outperforms TolerantECG in FMax by a clear margin (42.43\% vs 40.76\%), highlighting its superior precision-recall balance and robustness under class imbalance. Compared to other foundation models, including METS, MERL, and ECG-FM, both variants of \model consistently demonstrate substantial improvements.

These results confirm that incorporating attention supervision through CGAL does not compromise classification performance. On the contrary, it enables the model to learn more informative representations that translate into state-of-the-art results for downstream classification tasks.

Since \model-Rollout achieves the best localization performance, we use this variant for all subsequent experiments and analyses.

\subsection{Qualitative Analysis of Abnormality Localization}
Figure~\ref{fig:qualitative_compare} presents qualitative comparisons of abnormal region localization across multiple ECG conditions. For each example, we visualize the ground-truth abnormalities alongside the attention heatmaps generated by ECGAD$_\text{model}$ and \model. ECGAD$_\text{model}$ often produces sparse and fragmented activations, with limited alignment to the annotated abnormal segments. In contrast, \model generates more consistent and localized attention patterns that better overlap with the ground-truth regions.

Notably, \model demonstrates improved robustness across diverse conditions, including bundle branch blocks, myocardial infarction, and non-specific ischemic changes. The attention maps produced by \model are smoother and more temporally coherent, suggesting that the proposed CGAL mechanism effectively guides the model to focus on clinically relevant waveform regions. 
% These qualitative results complement the quantitative findings and highlight the advantage of supervision-guided attention in improving localization quality. Additional examples are provided in Figure~\ref{fig:qualitative_compare_sup} in the Supplementary material.

\section{Ablation Studies}
We conduct ablation experiments to  evaluate the contribution of each component in \model. To comprehensively assess both interpretability and predictive performance, we benchmark both abnormality localization and classification on $\text{ECGAD}_{\text{data}}$ dataset and additional classification on PTB-XL. This setup enables us to analyze how different design choices affect localization quality and classification accuracy under realistic clinical evaluation settings.

\begin{table}[t]
\centering
\caption{Effect of captioning loss on localization and classification.}
\label{tab:ablation_caption}
\resizebox{1.0\columnwidth}{!}{
\begin{tabular}{l|c|c|c}
\bf Configuration & \bf AUROC$_{\text{loc}}$ & \bf AUROC$_{\text{cls}}^{\text{ECGAD}}$ & \bf AUROC$_{\text{cls}}^{\text{PTB-XL}}$ \\
\midrule
$\beta = 1$ (\textbf{Default}) & \textbf{75.59 \tiny{$\pm$1.00}} & \underline{89.09 \tiny{$\pm$0.49}} & \textbf{92.04 \tiny{$\pm$0.29}} \\
$\beta = 2$ & \underline{75.28 \tiny{$\pm$0.95}} & \textbf{89.30 \tiny{$\pm$0.26}} & 91.36 \tiny{$\pm$0.13}\\
\end{tabular}
}
\end{table}

\textbf{Captioning.} 
Table~\ref{tab:ablation_caption} evaluates the effect of the captioning loss weight $\beta$ on localization and classification. The default setting, $\beta=1$, achieves the best localization performance and the highest PTB-XL classification performance, while remaining competitive on ECGAD classification. Increasing $\beta$ to 2 slightly improves ECGAD classification but leads to a small decline in both localization and PTB-XL classification performance. These results suggest that stronger captioning supervision does not consistently improve diagnostic performance and that $\beta=1$ provides a favorable balance between localization and classification.

\begin{table}
\centering
\caption{Effect of attention supervision on localization and classification.}
\label{tab:ablation_attn}
\resizebox{1.0\columnwidth}{!}{
\begin{tabular}{l|c|c|c}
\bf Configuration & \bf AUROC$_{\text{loc}}$ & \bf AUROC$_{\text{cls}}^{\text{ECGAD}}$ & \bf AUROC$_{\text{cls}}^{\text{PTB-XL}}$ \\
\midrule
$\gamma = 0$ & 61.33 \tiny{$\pm$1.27} & \textbf{89.41 \tiny{$\pm$0.23}} & \underline{92.14 \tiny{$\pm$0.09}} \\
$\gamma = 0.5$ (\textbf{Default}) & \textbf{75.55 \tiny{$\pm$1.00}} & 89.09 \tiny{$\pm$0.49} & 92.03 \tiny{$\pm$0.23} \\
$\gamma = 1.0$ & \underline{75.31 \tiny{$\pm$0.97}} & \underline{89.38 \tiny{$\pm$0.40}} & \textbf{92.61 \tiny{$\pm$0.12}} \\
\end{tabular}
}
\end{table}

\textbf{Attention Supervision.} Table~\ref{tab:ablation_attn} shows the effect of attention supervision controlled by $\gamma$. Removing attention guidance ($\gamma = 0$) maintains strong classification performance but leads to a sharp drop in localization, with AUROC$_{\text{loc}}$ decreasing to 61.33\%. Introducing strong attention supervision ($\gamma = 1.0$) substantially improves localization and yields the highest classification performance on PTB-XL (92.61\%). Our default setting ($\gamma = 0.5$) provides the most effective balance by achieving the highest localization score (75.55\%) while keeping competitive classification accuracy across both datasets.

% \begin{table}
% \centering
% \caption{Effect of Spatial-Temporal embedding. The Mean column reports the average of AUROC$_{\text{loc}}$ and AUROC$_{\text{cls}}$. \textcolor{red}{This is ablation on ST embedding module where Es, Et are the main contribution to this module. You need 4 different experiments including with/without each Es, Et. You do not need to train the model. Just run the interference: zero out feature that you do not use}}
% \label{tab:ablation_st}
% \resizebox{1.0\columnwidth}{!}{
%     \begin{tabular}{l|c|c|c}
%     \bf Configuration & \bf AUROC$_{\text{loc}}$ & \bf AUROC$_{\text{cls}}$ & \bf Mean \\
%     \midrule
%     w/o ST embedding ($\mathbf{E}^{s}_{l(n)} + \mathbf{E}^{t}_{t(n)}$) & 73.26 & \textbf{92.60} & 82.93 \\
%     w/ ST embedding (\bf \model) & \textbf{74.00} & 92.40 & \textbf{83.20} \\
%     \end{tabular}
% }
% \end{table}

% \textbf{ST Embedding.} Table~\ref{tab:ablation_st} shows the effect of incorporating spatial temporal (ST) embedding on both localization and classification performance. Removing ST embedding results in slightly higher classification performance, achieving the best AUROC$_{\text{cls}}$ of 92.60\%, but leads to reduced localization accuracy. In contrast, incorporating ST embedding improves localization performance to 74.00\%, while maintaining competitive classification results. This suggests that ST embedding provides useful structural information that enhances the model’s ability to capture spatial and temporal dependencies, which is particularly beneficial for abnormal region localization. 

\section{Conclusion}

We propose \textbf{\model}, an ECG foundation model that achieves strong diagnostic performance while enhancing clinically meaningful interpretability through abnormality localization, all without requiring manual attention annotations. \textbf{\model} is trained using a weakly-supervised mechanism that derives abnormality heatmap pseudo-labels from clinical heuristics. This approach guides the model's attention toward diagnostically relevant regions, effectively mitigating the localized annotation bottleneck in ECG data. By aggregating information from these clinically guided regions, \textbf{\model} improves the grounding of its predictions in relevant waveform features. Experiments demonstrate that \model delivers state-of-the-art performance on abnormality localization while maintaining robust classification capabilities.

\section*{Limitations}

% The CGAL pseudo-label generation relies on established clinical knowledge to identify abnormal regions during training. However, this approach may overlook certain cases due to diagnostic dependencies on patient-specific factors such as age and gender. The effectiveness of the method also depends on the quality and completeness of the pre-defined clinical rules and the CFR database. As a result, rare or unseen abnormalities not covered by these sources may not be properly detected. Although we partially mitigate these limitations by incorporating free-form text reports, some cases may still be missed.

% Additionally, while \model enhances interpretability by identifying waveform regions associated with its overall predictions, it is not a fully interpretable model. A single global CLS attention map cannot separately attribute specific segments to individual diagnoses in multi-diagnosis ECGs. Instead, the attention map highlights the union of all clinically relevant abnormal regions. This restricts the model from providing distinct, diagnosis-specific evidence for complex cases. 

% Future work may address these limitations by incorporating more comprehensive clinical guidelines to improve the quality and specificity of the generated pseudo-labels. In addition, future efforts could explore approaches to establish explicit mappings between localized waveform regions and individual diagnoses, enabling the model to distinguish diagnosis-specific evidence and offer more fine-grained localization.

First, CGAL's pseudo-label generation relies on predefined clinical rules and the CFR database. It may miss rare conditions or abnormalities tied to patient-specific factors (e.g., age, gender), though incorporating free-form reports partially mitigates this. 

Second, while \model enhances interpretability, it is not inherently fully interpretable. Its single global attention map highlights the union of all abnormal regions, preventing it from attributing specific waveform segments to individual diagnoses in multi-diagnosis ECGs. 

Future work will incorporate broader clinical guidelines to improve pseudo-label specificity and explore explicit mappings between localized regions and individual diagnoses to enable fine-grained, diagnosis-specific localization.

%Our approach has two main limitations. First, CGAL relies on predefined clinical rules and the CFR database, whose coverage may be incomplete and may not fully account for patient-specific factors or rare and unseen abnormalities. Although free-form reports partially mitigate this limitation, some cases may remain undetected. Second, the global CLS attention map localizes regions relevant to the overall prediction but cannot distinguish diagnosis-specific evidence in multi-diagnosis ECGs, limiting fine-grained interpretability. Future work will explore more comprehensive clinical knowledge and explicit mappings between waveform regions and individual diagnoses to improve pseudo-label quality and diagnosis-specific localization.

% \section*{Ethical considerations}
% %
% This work uses publicly available datasets (MIMIC-IV-ECG, ECGAD and PTB-XL) that have undergone institutional de-identification procedures. No personally identifiable information was accessed.  \model is intended to support clinical decision-making and is not a substitute for professional medical judgment. All predictions and recommendations must be reviewed and validated by qualified healthcare professionals before any clinical action is taken.

% \clearpage
{
    \small
    \bibliographystyle{ieeenat_fullname}
    \bibliography{main}
}

\clearpage
\setcounter{page}{1}
\appendix

\renewcommand{\thetable}{S\arabic{table}}
\renewcommand{\thefigure}{S\arabic{figure}}
\renewcommand{\theequation}{S\arabic{equation}}
\setcounter{table}{0} 
\setcounter{figure}{0}
\setcounter{equation}{0}

\twocolumn[
\begin{center}
    \Large \textbf{Supplementary Material: \model: Weakly-Supervised Interpretability-Enhancing Foundation Model for Electrocardiogram}
    \vspace{2em}
\end{center}
]
% \maketitlesupplementary
\section{CGAL Pseudo-Label Generation algorithm} \label{sec:appendix_cgal}

Algorithm~\ref{alg:cgal} outlines the procedure for curating pseudo‑heatmap labels that allocate diagnostic regions. The design ensures that pseudo‑labels are both clinically grounded and diagnosis‑specific. Before identifying abnormal regions, Stage 1 employs a pretrained UNet3+~\cite{unet3p, Joung2024DeepECGSegmentation} to segment the ECG signal into P wave, QRS complex, and T wave components. Stage 2 identifies candidate abnormal regions based on waveform segmentation and interval analysis, leveraging \textit{AddPseudoPwave} (Algorithm~\ref{alg:pseudoP}) to heuristically insert missing P‑wave segments and \textit{ComputeIntervals} (Algorithm~\ref{alg:compute_intervals}) to derive clinically relevant intervals including PR interval, QT interval, and ST segmenet. Stage 3 refines these candidate regions by intersecting them with diagnostic criteria retrieved from CFR, while also applying fallback rules (discussed in Section~\ref{sec:cfr_quality}) to enhance robustness against database incompleteness. As a result, the generated binary attention masks provide reliable supervision signals that guide the encoder to focus on clinically meaningful regions, thereby improving anomaly localization and interpretability in downstream tasks.

\begin{algorithm}
\caption{CGAL Pseudo-Label Generation}
\label{alg:cgal}
\begin{algorithmic}[1]
\REQUIRE ECG signal $\mathbf{X}$, Diagnosis list $\mathbf{D}$
\ENSURE Binary attention mask $\mathbf{m} \in \{0,1\}^N$
\STATE \textit{// Stage 1: Wave Segmenting}
\STATE $\mathbf{S} \gets \text{UNet3+}(\mathbf{X})$ \COMMENT{segment into P, QRS, T}

\STATE \textit{// Stage 2: Abnormal Seeking}
\STATE $(\mathbf{S}, \mathcal{R}) \gets \text{AddPseudoPwave}(\mathbf{S})$
\STATE $(\text{PR}, \text{QT}, \text{ST}) \gets \text{ComputeIntervals}(\mathbf{S})$
\STATE $\text{QTc} \gets \text{CorrectQT}(\text{QT}, \mathbf{S})$ 
% \COMMENT{Equation~\ref{eq:correct_qt}}
\STATE $\mathcal{R} \gets \mathcal{R} \cap \text{FlagAbnormal}(\text{S}, \text{PR}, \text{QT}, \text{QTc}, \text{ST})$

\STATE \textit{// Stage 3: Criteria Filtering}
\STATE $\mathbf{m} \gets \text{ZerosLike}(\mathbf{X})$
\FOR{$d$ in $\mathbf{D}$}
    \STATE $criteria \gets \text{CFR}(d)$
    \STATE $\mathcal{R} \gets \mathcal{R} \cap criteria$ \COMMENT{keep diagnosis-relevant}
    \STATE $\mathbf{m} \gets \mathbf{m} \lor \text{CreateMask}(\mathcal{R}, N)$
\ENDFOR
\RETURN $\mathbf{m}$
\end{algorithmic}
\end{algorithm}

\begin{algorithm}
\caption{AddPseudoPwave}
\label{alg:pseudoP}
\begin{algorithmic}[1]
\REQUIRE Segmentation mask $\mathbf{S}$
\ENSURE Modified segmentation $\mathbf{S}$, abnormal highlight set $\mathcal{R}$
\STATE $(\text{QRS}_{\text{on}}, \text{QRS}_{\text{off}}) \gets \text{GetOnOffset}(\mathbf{S} = 2)$
\STATE $(\text{T}_{\text{on}}, \text{T}_{\text{off}}) \gets \text{GetOnOffset}(\mathbf{S} = 3)$
\FOR{$i \in \{1, \ldots, len(\text{QRS}_{\text{on}})\}$}
    \STATE $\text{start} \gets \text{T}_{\text{off}}[i-1]$
    \STATE $\text{end} \gets \text{QRS}_{\text{on}}[i]$
    \IF{$1 \notin \mathbf{S}[\text{start}:\text{end}]$}
        \STATE $\text{area} \gets \text{end} - \text{start}$
        \STATE $\text{startP} \gets \text{start} + \frac{\text{area}}{3}$
        \STATE $\text{endP} \gets \text{end} - \frac{\text{area}}{4}$
        \STATE $\mathbf{S}[\text{startP}:\text{endP}] \gets 1$ \COMMENT{Mark pseudo‑P wave region}
        \STATE $\mathcal{R} \gets \mathcal{R} \cup \text{Flag}(\text{startP}, \text{endP})$
    \ENDIF
\ENDFOR
\RETURN $\mathbf{S}, \mathcal{R}$
\end{algorithmic}
\end{algorithm}

\begin{algorithm}
\caption{ComputeIntervals}
\label{alg:compute_intervals}
\begin{algorithmic}[1]
\REQUIRE Segmentation mask $\mathbf{S}$
\ENSURE PR interval, QT interval and ST segment
\STATE \textit{// Step 1: Detect onset/offset indices}
\STATE $(\text{P}_{\text{on}}, \text{P}_{\text{off}}) \gets \text{GetOnOffset}(\mathbf{S} = 1)$
\STATE $(\text{QRS}_{\text{on}}, \text{QRS}_{\text{off}}) \gets \text{GetOnOffset}(\mathbf{S} = 2)$
\STATE $(\text{T}_{\text{on}},\text{T}_{\text{off}}) \gets \text{GetOnOffset}(\mathbf{S} = 3)$
\STATE \textit{// Step 2: Initialize empty intervals}
\STATE $\text{PR} \gets \text{zeros}(\text{len}(\mathbf{S}))$
\STATE $\text{QT} \gets \text{zeros}(\text{len}(\mathbf{S}))$
\STATE $\text{ST} \gets \text{zeros}(\text{len}(\mathbf{S}))$
\STATE \textit{// Step 3: Construct intervals}
\FOR{each $(\text{start}, \text{end})$ in $(\text{P}_{\text{on}}, \text{QRS}_{\text{on}})$}
    \STATE $\text{PR}[\text{start}:\text{end}] \gets 1$
\ENDFOR
\FOR{each $(\text{start}, \text{end})$ in $(\text{QRS}_{\text{on}}, \text{T}_{\text{off}})$}
    \STATE $\text{QT}[\text{start}:\text{end}] \gets 1$
\ENDFOR
\FOR{each $(\text{start}, \text{end})$ in $(\text{QRS}_{\text{off}}, T_{\text{on}})$}
    \STATE $\text{ST}[\text{start}:\text{end}] \gets 1$
\ENDFOR
\RETURN PR, QT, ST
\end{algorithmic}
\end{algorithm}

\section{UNet3+ performance} \label{sec:unet_performance}

Accurate segmentation is a critical prerequisite for reliable ECG analysis, as it enables precise identification of individual wave components within the input signal. Inadequate performance at this stage can propagate errors into subsequent modules, leading to incorrect pseudo-heatmap label curation and causing the encoder to attend to partially erroneous regions. Table~\ref{tab:segment_perform} summarizes the comparative performance of three segmentation approaches, namely ProMac, NeuroKit, and UNet3+, in detecting P, QRS, and T wave boundaries. Evaluation was conducted on the Semi-Seg-ECG test set \cite{semisegecg} using the F1-score metric \cite{Joung2024DeepECGSegmentation}, which provides a balanced measure of precision and recall for boundary localization.
\begin{table}
\centering
\caption{Performance comparison of segmentation methods on ECG component boundary localization, evaluated using F1‑score on the Semi‑Seg‑ECG test set.}
\resizebox{1.0\columnwidth}{!}{
\begin{tabular}{l|c|c|c|c|c|c}
\bf Method & \bf P onset & \bf P offset & \bf QRS onset & \bf QRS offset & \bf T onset & \bf T offset \\
\midrule
ProMac    & 84.72 & 84.91 & 77.05 & 78.08 & 81.70 & 67.62 \\
NeuroKit  & 84.78 & 84.60 & 79.62 & 80.92 & 79.52 & 63.97 \\
UNet3+    & \textbf{96.68} & \textbf{96.75} & \textbf{99.88} & \textbf{99.89} & \textbf{97.82} & \textbf{96.66} \\
\end{tabular}
}
\label{tab:segment_perform}
\end{table}

Across all wave boundaries, UNet3+ demonstrates substantially higher accuracy, achieving near‑perfect F1‑scores for QRS onset and offset (99.88\% and 99.89\%, respectively). In contrast, ProMac and NeuroKit exhibit moderate performance, particularly struggling with T‑wave offset detection (67.62\% and 63.97\%). These results highlight the importance of robust segmentation: errors in boundary detection disproportionately affect downstream anomaly localization, as misaligned pseudo‑labels can misguide the encoder’s attention. The superior performance of UNet3+ underscores its suitability as the backbone segmentation model in our pipeline, ensuring reliable wave delineation and minimizing error propagation in subsequent stages.

\begin{table}
\centering
\caption{F1 score performance of the UNet3+ across multiple datasets in the SemiSegECG test set.}
\label{tab:segmentation}
\resizebox{1.0\columnwidth}{!}{
\begin{tabular}{l|c|c|c|c|c|c}
\bf Dataset & \bf P onset & \bf P offset & \bf QRS onset & \bf QRS offset & \bf T onset & \bf T offset \\
\midrule
ISP        & 97.09 & 97.16 & 100.00 & 100.00 & 99.50 & 99.48 \\
Zhejiang   & 94.56 & 94.60 & 99.84  & 99.85  & 96.92 & 95.73 \\
LUDB       & 96.49 & 96.57 & 99.86  & 99.86  & 99.60 & 99.27 \\
QTDB       & 97.74 & 97.74 & 99.43  & 99.43  & 95.52 & 89.14 \\
\end{tabular}
}
\end{table}

For a more detailed breakdown, table~\ref{tab:segmentation} presents the F1 score performance of the segmentation model across multiple datasets within the SemiSegECG test set. The model achieves consistently high accuracy across all waveform components, with near perfect performance on QRS onset and offset detection across all datasets. P wave segmentation shows slightly lower but still strong results, particularly on the Zhejiang dataset. T wave segmentation exhibits greater variability, with noticeable performance drops on QTDB, especially for T wave offset. Overall, these results demonstrate the robustness of the segmentation model across diverse datasets, while highlighting the increased difficulty of accurately detecting subtle waveform boundaries.

\section{Abnormal criteria in ECG signal} \label{sec:abnormal_threshold}
In the second stage of the disease heatmap curation pipeline, we identify and highlight abnormal regions across individual waves and leads. To achieve this, we collect diagnostic conditions from LITFL~\cite{litfl} and subsequently verify them with clinical experts to ensure medical validity. These curated conditions serve as the foundation for constructing reliable heatmaps that emphasize clinically relevant abnormalities. A summary of the identified conditions is presented in Table~\ref{tab:abnormal_criteria}.
\begin{table}
\centering
\caption{Abnormal criteria for each ECG waveform component. These thresholds are used in CGAL Stage~2 to flag candidate abnormal regions.}
\label{tab:abnormal_criteria}
\resizebox{1.0\columnwidth}{!}{
\begin{tabular}{ll}
\textbf{Component} & \textbf{Abnormal Conditions} \\
\midrule
\multirow{5}{*}{P wave} 
    & Missing (absent P wave) \\
    & Amplitude $> 0.25$ mV in limb leads \\
    & Amplitude $> 0.15$ mV in precordial leads \\
    & Duration $> 120$ ms \\
    & Inverted in leads I, II, III; Upright in lead aVR \\
\midrule
\multirow{2}{*}{QRS complex} 
    & Duration $> 120$ ms \\
    & Abnormal morphology (RSR', QR, rS patterns) \\
\midrule
\multirow{3}{*}{T wave} 
    & Amplitude $> 0.5$ mV in limb leads \\
    & Amplitude $> 1.0$ mV in precordial leads \\
    & Inverted in normally upright leads (I, II, V2--V6) \\
\midrule
\multirow{2}{*}{PR interval} 
    & Duration $> 200$ ms (prolonged) \\
    & Duration $< 120$ ms (shortened) \\
\midrule
\multirow{2}{*}{QT interval} 
    & Corrected duration $> 440$ ms (prolonged) \\
    & Corrected duration $< 350$ ms (shortened) \\
\midrule
\multirow{2}{*}{ST segment} 
    & Elevation $> 0.1$ mV ($> 0.05$ mV in V2, V3) \\
    & Depression $< -0.5$ mV from isoelectric line \\
\end{tabular}
}
\end{table}

\section{CFR quality and fallback} \label{sec:cfr_quality}

The CFR component is designed to retrieve clinically relevant diagnostic criteria. To evaluate its performance, we collect all diagnostic classes from the MIMIC-IV-ECG dataset~\cite{mimic-iv} and use them as query inputs to measure the acceptance rate. The results show an acceptance rate of 20.18\%, with an average similarity score of 84.95 among the accepted queries. This indicates that, while the retrieved criteria are generally of high quality when matched, the overall retrieval coverage remains limited.

Analysis of the retrieval errors reveals that many query reports already describe clinical criteria rather than diagnostic labels, and therefore do not directly correspond to entries in the CFR database. Consequently, such inputs cannot be effectively matched to diagnosis–criteria pairs. Table~\ref{tab:cfr_errors} provides examples of low scoring queries, where clinically valid descriptions such as ``nonspecific ST changes'' or ``prolonged PR interval'' are not retrieved. 

\begin{table}
\centering
\caption{Examples of lowest‑scoring queries in CFR retrieval.}
\resizebox{1.0\columnwidth}{!}{
\begin{tabular}{l|l|c}
\bf Query & \bf Retrieved diagnosis & \bf Retrieve score \\
\midrule
Nonspecific ST changes & Septal hypertrophy & -14.26 \\
Marked ST depression in V lead & Pericarditis & -10.04 \\
Short PR interval & Extreme Axis Deviation & -9.29 \\
ST depression V1-V3 & Poor R Wave Progression & -8.83 \\
Prolonged PR interval & R Wave Peak Time & -8.04 \\
ST-T changes & Digitalis Effect & -7.53 \\
rSr' pattern in V1 or V2 & Low QRS voltage & -4.20 \\
Inferior nonspecific ST-T changes & Septal hypertrophy & -3.83 \\
Abnormally tall P in V1 or V2 & Raised Intracranial Pressure & -3.31 \\
Nonspecific lateral ST-T changes & Benign Early Repolarisation & -2.43 \\
\end{tabular}
}
\label{tab:cfr_errors}
\end{table}

To mitigate these limitations, we adopt a fallback strategy: when diagnostic features cannot be retrieved, the diagnosis name itself is used to guide abnormal heatmap filtering. Specifically, if a query explicitly mentions a wave component (e.g., “PR interval,” “T‑waves”), we retain the corresponding highlights from Stage~2 of CGAL. For broader diagnostic terms such as “ischaemia,” “ischemic,” or “infarction,” we emphasize the ST segment and T wave, while terms like “atrial” direct attention to the P wave. In addition, ECG lead group phrases (e.g., anterior, septal, inferior) are leveraged to ensure that abnormal heatmap labels highlight the correct anatomical regions. This strategy helps reduces false‑negative annotations and produces more accurate pseudo‑labels for attention supervision.

\section{Ablation on Spatial-Temporal Embedding} \label{sec:ablation_st}

To evaluate the contribution of the Spatial-Temporal (ST) embedding (detailed in Section~\ref{sec:st-embed} of the main text), we compare our default model against a variant trained without the ST embedding components ($\mathbf{E}^{s}$ and $\mathbf{E}^{t}$). The ST embedding provides explicit lead identity and time frame information, enriching each token with spatial and temporal context.

Table~\ref{tab:ablation_st} summarizes the localization and classification performance of both configurations. 

\begin{table}[h]
\centering
\caption{Effect of Spatial-Temporal embedding on localization and classification.}
\label{tab:ablation_st}
\resizebox{1.0\columnwidth}{!}{
    \begin{tabular}{l|c|c|c}
    \bf Configuration & \bf AUROC$_{\text{loc}}$ & \bf AUROC$_{\text{cls}}^{\text{ECGAD}}$ & \bf AUROC$_{\text{cls}}^{\text{PTB-XL}}$ \\
    \midrule
    w/o ST embedding & 75.01 \tiny{$\pm$0.84} & \textbf{90.31 \tiny{$\pm$0.24}} & \textbf{92.08 \tiny{$\pm$0.30}} \\
    w/ ST embedding (\textbf{Default}) & \textbf{75.55 \tiny{$\pm$1.00}} & 89.09 \tiny{$\pm$0.49} & 92.03 \tiny{$\pm$0.23} \\
    \end{tabular}
}
\end{table}

Removing the ST embedding results in slightly higher classification performance, achieving an AUROC$_{\text{cls}}^{\text{ECGAD}}$ of 90.31\% and an AUROC$_{\text{cls}}^{\text{PTB-XL}}$ of 92.08\%. However, this omission leads to reduced localization accuracy (75.01\%). In contrast, incorporating the ST embedding improves localization performance to 75.55\%, while still maintaining competitive classification results. This suggests that the ST embedding provides useful structural information that enhances the model’s ability to capture spatial and temporal dependencies across leads. Such explicit spatial-temporal grounding proves particularly beneficial for precisely identifying localized abnormalities, confirming its value in optimizing the model for clinical interpretability.

\section{Qualitative comparison}

\begin{figure*}[h]
    \centering
    \includegraphics[width=1.0\linewidth]{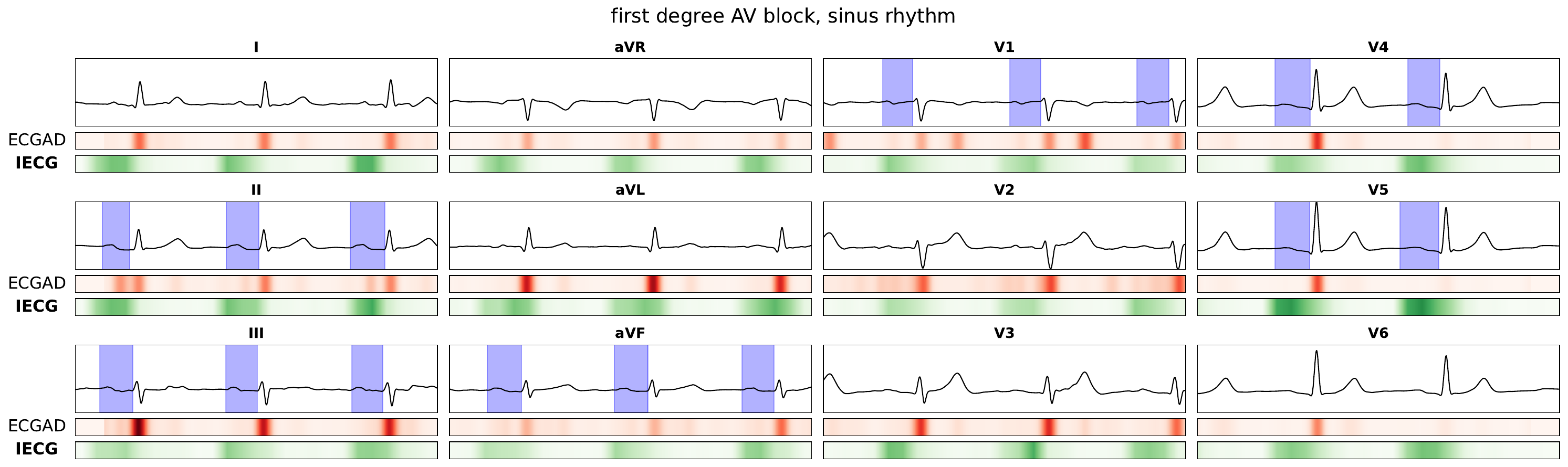}
    \caption{First-degree atrioventricular block with sinus rhythm. \model consistently highlights prolonged PR intervals across multiple leads, showing stronger alignment with ground truth compared to the baseline, which exhibits more dispersed and less structured localization.}
    \label{fig:qual_1}
\end{figure*}

\begin{figure*}[h]
    \centering
    \includegraphics[width=1.0\linewidth]{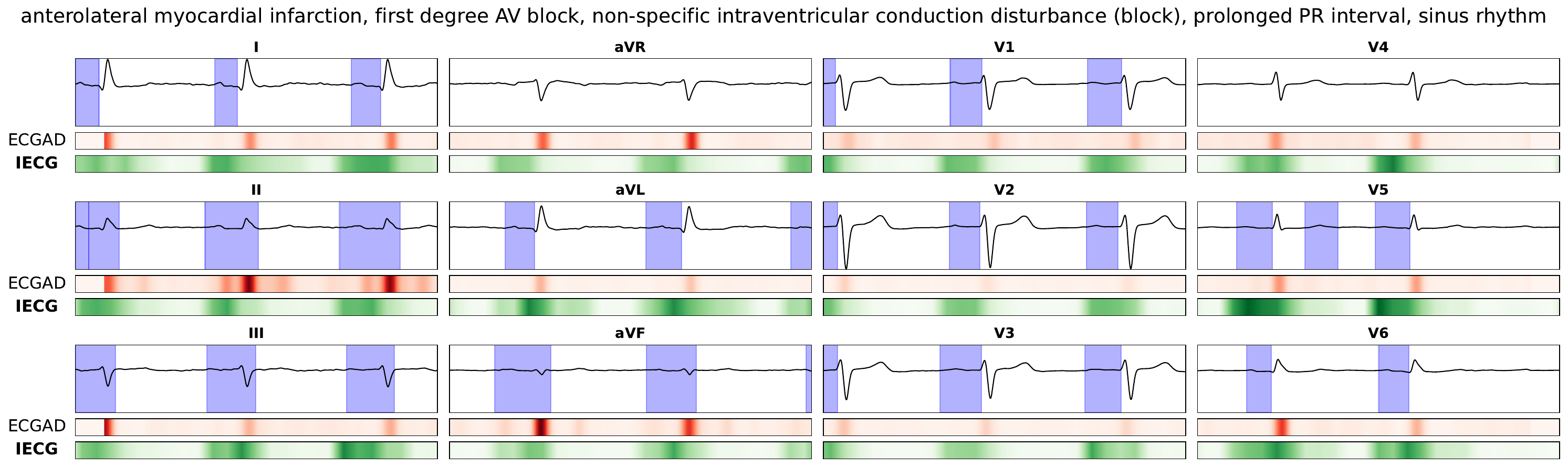}
    \caption{Anterolateral myocardial infarction with first-degree AV block and intraventricular conduction disturbance. As expected clinically, \model assigns substantially higher attention weight to the anterolateral leads (I, aVL, and V4--V6) to capture the myocardial infarction, while simultaneously maintaining strong focus on the prolonged PR intervals across the waveform to identify the AV block. This demonstrates the model's ability to accurately localize both spatially-restricted morphological abnormalities and global temporal conduction delays within the same sample, whereas the baseline exhibits fragmented and less anatomically specific activations.}
    \label{fig:qual_2}
\end{figure*}

\begin{figure*}[h]
    \centering
    \includegraphics[width=1.0\linewidth]{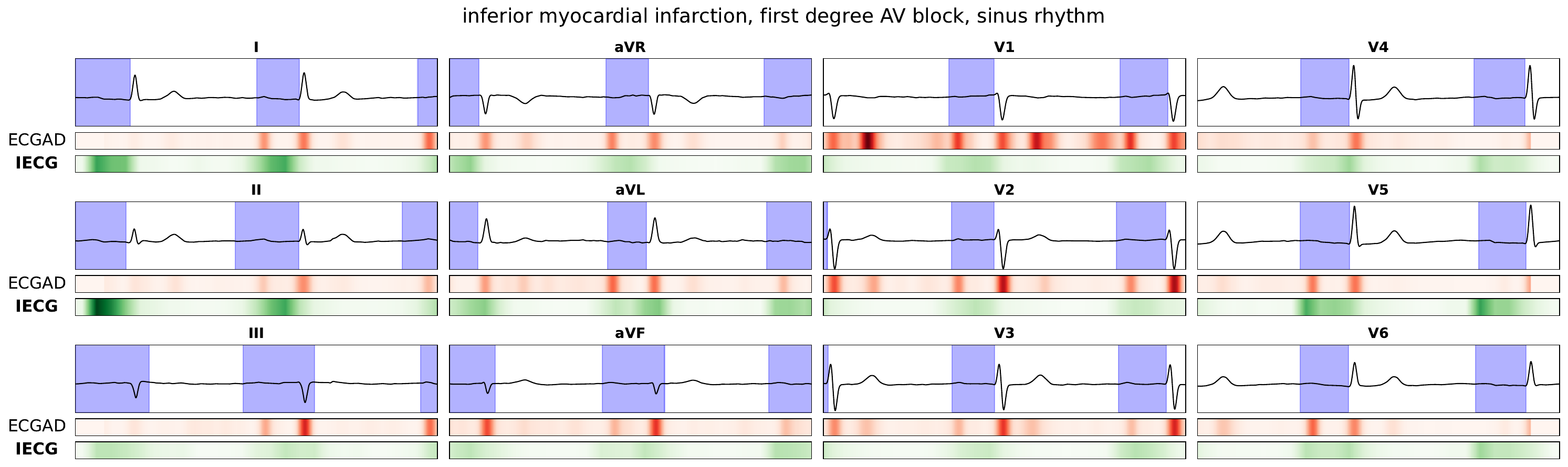}
    \caption{Inferior myocardial infarction with first-degree AV block. To accurately diagnose the inferior infarction, \model distinctly concentrates its attention on the inferior leads (II), precisely matching the expected anatomical region of the ischemic injury. Concurrently, it correctly highlights the prolonged PR intervals to account for the atrioventricular conduction delay. In contrast, the baseline model fails to prioritize the inferior leads and produces activations that lack clear anatomical or temporal justification.}
    \label{fig:qual_3}
\end{figure*}

\begin{figure*}[h]
    \centering
    \includegraphics[width=1.0\linewidth]{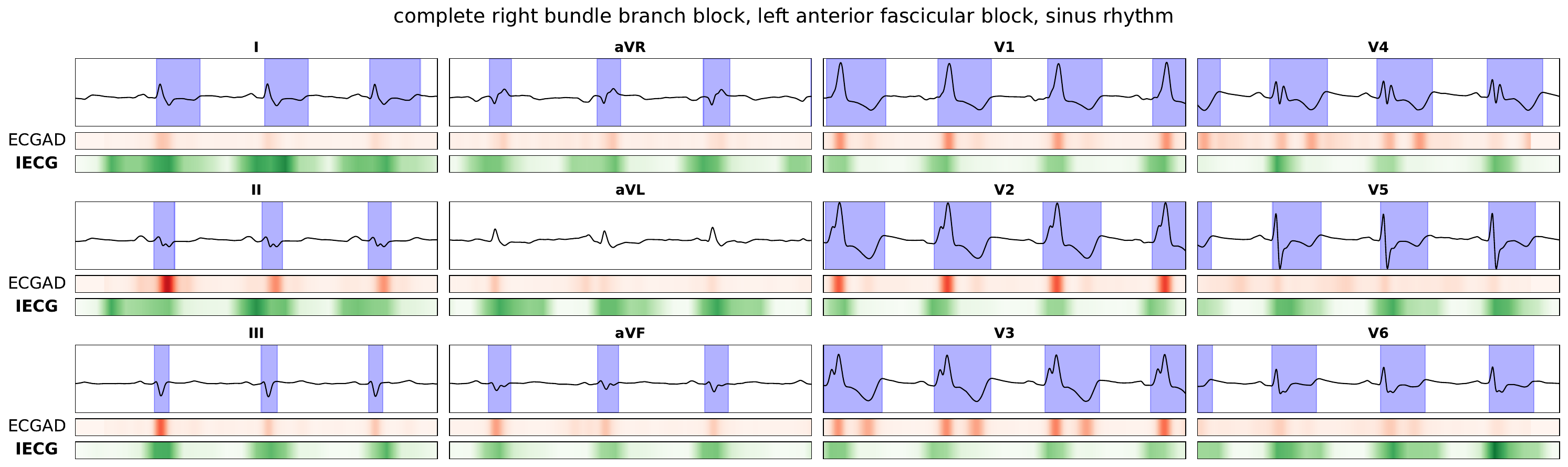}
    \caption{Complete right bundle branch block (RBBB) with left anterior fascicular block (LAFB). \model accurately targets the broadened QRS complexes characteristic of RBBB, placing prominent attention on the right precordial leads (V1--V3). Furthermore, to identify LAFB, the model correctly distributes its focus toward the inferior leads (II, III, aVF) and high lateral leads (I, aVL), which are critical for recognizing the associated left axis deviation. This precise, anatomically-grounded localization contrasts with the baseline, which generates uncoordinated activations that fail to clearly reflect the underlying multi-fascicular conduction pathologies.}
    \label{fig:qual_4}
\end{figure*}

To further assess the quality of abnormal region localization, we provide visual comparisons across multiple ECG samples with diverse diagnostic conditions, as shown in Figures~\ref{fig:qual_1} to~\ref{fig:qual_4}. These examples illustrate how well the predicted attention maps align with expert annotations across both temporal and lead dimensions. Each example shows a 12-lead ECG signal with corresponding localization results. For each lead, the waveform is shown together with ground truth annotations (blue), baseline predictions from $\text{ECGAD}_{\text{model}}$ (red), and predictions from \model (green). Rather than relying solely on aggregate metrics, this analysis offers a more detailed view of model behavior, highlighting strengths and failure modes in complex and multi-label scenarios. 

\section{Limitations of Ground-Truth Abnormal Annotations}

\begin{figure}[t]
  \centering
    \includegraphics[width=1.0\linewidth]{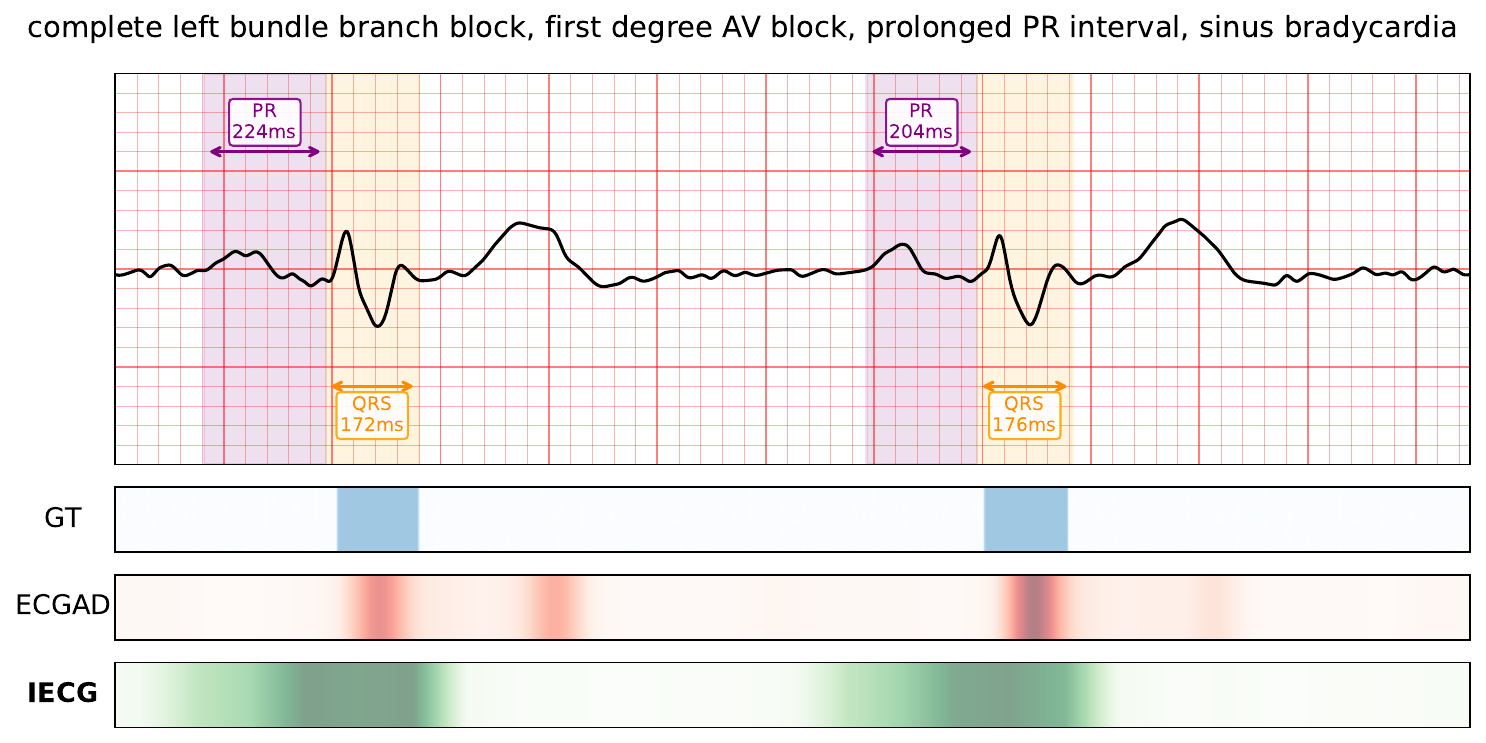}
    \caption{Example illustrating annotation incompleteness in $\text{ECGAD}_{\text{data}}$ using Lead II. The ECG exhibits complete left bundle branch block (LBBB) and first-degree atrioventricular (AV) block, as indicated by prolonged QRS duration (e.g., 172--176 ms) and prolonged PR interval (e.g., 204--224 ms), respectively. The ground-truth (GT) heatmap highlights regions corresponding to widened QRS complexes associated with LBBB but fails to capture the prolonged PR intervals that define first-degree AV block. In contrast, \model accurately highlights both QRS and PR regions, reflecting clinically relevant abnormalities across both morphological (ventricular depolarization) and temporal (atrioventricular conduction delay) features. This example demonstrates that missing annotations of temporal abnormalities can lead to underestimation of localization performance.}
    \label{fig:ecgad_error}
\end{figure}

Although \model achieves strong localization performance, the overall AUROC$_\text{loc}$ could be constrained by inaccuracies and incompleteness in the ground-truth annotations of $\text{ECGAD}_{\text{data}}$. In particular, the provided abnormal region labels do not always capture all clinically relevant waveform features, especially for conditions characterized by temporal abnormalities rather than distinct morphological changes.

A representative example is shown in Figure~\ref{fig:ecgad_error}, where the ECG exhibits complete left bundle branch block (LBBB) and first-degree atrioventricular (AV) block. The annotations correctly highlight the prolonged QRS complexes associated with LBBB, which is defined by a widened QRS duration ($\geq 120$ ms) reflecting delayed ventricular depolarization~\cite{lbbb_litfl}. However, they fail to capture the prolonged PR interval ($>200$ ms) that characterizes first-degree AV block, which corresponds to delayed conduction between atrial and ventricular activation~\cite{avblock_litfl, avblock_acls}.

Clinically, these abnormalities manifest in distinct regions of the waveform: LBBB primarily affects the QRS complex, while first-degree AV block is defined by prolongation of the PR interval preceding the QRS. As a result, the ground-truth heatmaps, which focus on QRS-based abnormalities, omit diagnostically important temporal features in the PR segment. 

In contrast, \model highlights both the widened QRS complexes and the prolonged PR intervals, producing attention maps that more closely reflect the full set of clinically relevant abnormalities. Such discrepancies indicate that the ground-truth annotations do not fully capture all relevant diagnostic evidence, causing the model to be penalized for correctly identifying waveform regions that are not labeled. This limitation partially explains the gap between the strong qualitative alignment of \model and its quantitative evaluation scores.

\section{Evaluating the Effectiveness of CGAL Pseudo-labels}

To further understand the contribution of the CGAL pipeline, we isolate its performance by evaluating the generated pseudo-labels directly against the expert ground-truth annotations in the ECGAD dataset. In this experiment, referred to as \textit{Just CGAL}, we construct abnormal heatmaps exclusively using the rule-based clinical heuristics and CFR module (as detailed in Section~\ref{sec:cgal} of the main text), entirely bypassing the Transformer model's attention rollout. 

We compare this heuristic-only baseline against the learned representations of our model, both with and without attention supervision. Table~\ref{tab:cgal_effectiveness} presents the localization performance (AUROC$_{\text{loc}}$) for each configuration.

\begin{table}[h]
\centering
\caption{Comparison of localization performance between the heuristic-based CGAL pseudo-labels and the trained models.}
\label{tab:cgal_effectiveness}
\resizebox{1.0\columnwidth}{!}{
    \begin{tabular}{l|c}
    \bf Method & \bf AUROC$_{\text{loc}}$ \\
    \midrule
    \model w/o Attention Supervision ($\gamma = 0$) & 61.33\% \\
    Just CGAL (Rule-based Pseudo-labels) & 70.13\% \\
    \model ($\gamma = 1.0$) & 75.31\% \\
    \model (\textbf{Default}, $\gamma = 0.5$) & \textbf{75.55\%} \\
    \end{tabular}
}
\end{table}

The results yield two key observations. First, the heuristic-derived heatmaps from \textit{Just CGAL} achieve an AUROC$_{\text{loc}}$ of 70.13\%, significantly outperforming the baseline \model trained without attention supervision (61.33\%). This validates the quality and clinical relevance of the generated pseudo-labels. Second, and most importantly, when these pseudo-labels are used to supervise the model during training, the resulting models consistently surpass the performance of their own pseudo-label ``teacher,'' achieving 75.31\% ($\gamma = 1.0$) and 75.55\% ($\gamma = 0.5$). This demonstrates that the weakly-supervised learning process allows the model to generalize beyond the rigid, rule-based heuristics of CGAL, learning a more robust and comprehensive understanding of morphological abnormalities from large-scale data regardless of the exact supervision weight.

\section{Report Generation}

To evaluate the capabilities of our learned ECG representations on complex generative tasks, we assess \model on ECG report generation. Following the implementation of ECG-Chat~\cite{ecg-chat}, we utilize a Large Language Model (LLM) initialized with Vicuna-13B (v1.5) weights and align our frozen ECG encoder to the LLM using a projection layer. We evaluate the generated reports using standard natural language generation metrics, including BLEU-4, ROUGE-L, METEOR, and BERT-score (F1). 

To understand the impact of CGAL on generative performance, we compare our default \model with a variant trained without attention supervision (w/o CGAL). Both models are evaluated on the PTB-XL test set, using the corresponding English-translated reports as references. The results are summarized in Table~\ref{tab:report_generation}.

\begin{table}[h]
\centering
\caption{Report generation performance on the PTB-XL dataset.}
\label{tab:report_generation}
\resizebox{1.0\columnwidth}{!}{
\begin{tabular}{l|c|c|c|c}
\bf Configuration & \bf BLEU-4 & \bf ROUGE-L & \bf METEOR & \bf BERT-score \\
\midrule
w/o CGAL ($\gamma = 0$) & \textbf{12.57} & \textbf{34.49} & \textbf{33.61} & \textbf{89.25} \\
\model (\textbf{Default}) & 12.14 & 34.28 & 32.56 & 89.20 \\
\end{tabular}
}
\end{table}

As shown in Table~\ref{tab:report_generation}, the model trained without CGAL achieves slightly higher scores across traditional n-gram based metrics such as BLEU-4 and METEOR. However, the performance of the default \model remains highly competitive, demonstrating identical semantic similarity as measured by BERT-Score (89.20\%) and only a marginal decrease in ROUGE-L (34.28 vs 34.49). These findings indicate that while explicitly constraining the model's attention to clinically relevant regions via CGAL imposes a slight regularization effect on exact text matching, it preserves the rich semantic representations necessary for accurate and coherent clinical report generation.

\section{Classification on CPSC and Chapman Datasets}

To further demonstrate the robustness and generalization ability of our learned representations, we evaluate \model on two additional downstream classification benchmarks: CPSC 2018~\cite{cpsc} and Chapman~\cite{chapman}. For fair comparison, we adopt the same data splitting strategy for these datasets as established in MERL~\cite{merl}.

Table~\ref{tab:cpsc_chapman_classify} presents the classification results. Consistent with our findings on PTB-XL, \model achieves highly competitive performance across both datasets, outperforming all baselines except TolerantECG, which it trails by a small margin. Specifically, on the CPSC dataset, \model achieves an AUROC of 96.32\% and an FMax of 80.06\%, while on the Chapman dataset, it achieves an AUROC of 95.32\% and an FMax of 52.15\%. These results confirm that incorporating attention supervision through CGAL for interpretability does not significantly compromise the model's diagnostic classification capabilities, enabling it to learn informative features that generalize well to diverse unseen datasets.

\begin{table}[h]
    \centering
    \caption{Classification results evaluated on the CPSC and Chapman datasets. \textbf{Bold} indicates the best result, and \underline{underline} indicates the second-best result.}
    \label{tab:cpsc_chapman_classify}
    \resizebox{1.0\columnwidth}{!}{
        \begin{tabular}{l|c|c|c|c}
        \multirow{2}{*}{\bf Model} & \multicolumn{2}{c|}{\bf CPSC} & \multicolumn{2}{c}{\bf Chapman} \\
        \cmidrule{2-5}
        & \bf FMax & \bf AUROC & \bf FMax & \bf AUROC \\
        \midrule
        ST-MEM \cite{stmem} & 26.95 \tiny{$\pm$0.09} & 63.03 \tiny{$\pm$0.11} & 12.47 \tiny{$\pm$0.32} & 62.22 \tiny{$\pm$0.80} \\
        HeartLang \cite{jin2025readingheartlearningecg} & 36.44 \tiny{$\pm$0.61} & 70.13 \tiny{$\pm$0.45} & 29.52 \tiny{$\pm$0.65} & 84.15 \tiny{$\pm$0.64} \\
        XAI4ECG \cite{xai4ecg} & 57.46 \tiny{$\pm$0.65} & 85.74 \tiny{$\pm$0.40} & 29.21 \tiny{$\pm$0.78} & 82.10 \tiny{$\pm$0.58} \\
        ECG-FM \cite{ecgfm} & 64.94 \tiny{$\pm$0.38} & 90.24 \tiny{$\pm$0.12} & 44.03 \tiny{$\pm$0.71} & 90.76 \tiny{$\pm$0.12} \\
        ECGAD \cite{ecgad} & 68.89 \tiny{$\pm$0.90} & 91.59 \tiny{$\pm$0.82} & 34.41 \tiny{$\pm$0.73} & 87.88 \tiny{$\pm$0.38} \\
        MERL \cite{merl} & 77.98 \tiny{$\pm$0.17} & 95.51 \tiny{$\pm$0.06} & 50.43 \tiny{$\pm$0.20} & 94.69 \tiny{$\pm$0.12} \\
        METS \cite{mets} & 79.16 \tiny{$\pm$0.41} & 96.12 \tiny{$\pm$0.07} & 49.67 \tiny{$\pm$0.21} & 95.08 \tiny{$\pm$0.05} \\
        TolerantECG \cite{tolerantecg} & \textbf{81.53 \tiny{$\pm$0.40}} & \textbf{96.73 \tiny{$\pm$0.05}} & \textbf{53.70 \tiny{$\pm$0.23}} & \textbf{95.88 \tiny{$\pm$0.03}} \\
        \midrule
        \model (Ours) & \underline{80.06 \tiny{$\pm$0.36}} & \underline{96.32 \tiny{$\pm$0.07}} & \underline{52.15 \tiny{$\pm$0.36}} & \underline{95.32 \tiny{$\pm$0.05}} \\
        \end{tabular}
    }
\end{table}

\section{Trade-off Between Interpretability and Downstream Performance}

A central challenge in developing medical AI systems is balancing raw predictive performance with clinical interpretability. Across the experiments detailed in the main text and this supplementary material, we consistently observe a highly favorable trade-off when incorporating CGAL.

\textbf{Impact on Classification:} As demonstrated in the main text's ablation studies, training \model without attention supervision ($\gamma = 0$) maximizes classification metrics, achieving an AUROC$_{\text{cls}}$ of 92.42\% on PTB-XL and 90.87\% on ECGAD. However, this comes at the severe cost of interpretability, with localization performance dropping to 61.33\%. By introducing CGAL with $\gamma = 0.5$, the model gains a massive +14.22\% absolute improvement in localization (AUROC$_{\text{loc}}$ of 75.55\%). This structural regularization imposes only a marginal penalty on diagnostic power: AUROC$_{\text{cls}}$ remains highly competitive at 92.21\% on PTB-XL and 89.09\% on ECGAD. Furthermore, the robust generalization of this balanced representation is confirmed on external benchmarks, where \model maintains second-best overall performance on both the CPSC (96.32\%) and Chapman (95.32\%) datasets.

\textbf{Impact on Report Generation:} A similar regularizing effect is observed in complex generative tasks. As shown in the previous section on Report Generation, removing CGAL yields slightly higher exact text-matching scores (e.g., BLEU-4 of 12.57\% vs. 12.14\%). Nonetheless, the default \model retains near-identical semantic coherence (BERT-score of 89.20\% vs. 89.25\%). 

\textbf{Architectural Contributions to the Trade-off:} This favorable balance is fundamentally enabled by our specific architectural design choices, namely Attention Rollout and Spatial-Temporal (ST) embedding. As demonstrated in the main text's experiments, applying attention supervision solely to the last attention layer yields a slightly higher classification AUROC$_{\text{cls}}$ on ECGAD (90.03\%) compared to Attention Rollout (89.09\%), but results in reduced localization accuracy (73.94\% vs. 75.55\%). Similarly, as detailed in Section~\ref{sec:ablation_st}, removing the ST embedding marginally improves classification (90.31\% on ECGAD) but degrades localization (75.01\% vs. 75.55\%). Both Attention Rollout and ST embedding introduce structural priors that regularize the network, intentionally trading a small fraction of raw predictive capacity to achieve more precise and clinically interpretable spatial-temporal grounding.

\textbf{Conclusion:} These results indicate that explicitly constraining the model's attention to align with clinical heuristics inherently acts as an anatomical and temporal regularizer. While this regularization slightly bounds the unconstrained optimization of downstream objectives, the resulting trade-off is highly advantageous. \model secures state-of-the-art or highly competitive predictive performance across diverse tasks while guaranteeing that its decisions are grounded in clinically valid waveform evidence, which is a critical prerequisite for clinical trust and real-world medical deployment.

\section{Ethical considerations}
This work uses publicly available datasets (MIMIC-IV-ECG, PTB-XL, ECGAD, CPSC and Chapman) that have undergone institutional de-identification procedures. No personally identifiable information was accessed.  \model is intended to support clinical decision-making and is not a substitute for professional medical judgment. All predictions and recommendations must be reviewed and validated by qualified healthcare professionals before any clinical action is taken.

\end{document}